\documentclass[10pt,twocolumn,letterpaper]{article}

\usepackage{wacv}              

\usepackage{graphicx}
\usepackage{amsmath}
\usepackage{amssymb}
\usepackage{booktabs}

\usepackage[pagebackref,breaklinks,colorlinks]{hyperref}

\usepackage[capitalize]{cleveref}
\crefname{section}{Sec.}{Secs.}
\Crefname{section}{Section}{Sections}
\Crefname{table}{Table}{Tables}
\crefname{table}{Tab.}{Tabs.}

\def\wacvPaperID{369} 
\def\confName{WACV}
\def\confYear{2025}

\newcommand{\plad}{PLAD}
\usepackage{bbm, bm}
\usepackage{stmaryrd}
\usepackage{standalone}
\usepackage{amsmath}
\newcommand{\stdv}[1]{\scriptsize$\pm$#1}
\newcommand{\fset}[1]{\mathcal{#1}}
\newcommand{\ffield}[1]{\mathbb{#1}}
\newcommand{\ffunc}[1]{\mathfrak{#1}}
\newcommand{\fdim}[1]{#1}

\newcommand{\fvm}[1]{\mathbf{#1}}
\newcommand{\fgrvm}[1]{\boldsymbol{#1}}
\newcommand{\ours}{PLUME} 
\usepackage{tikz}
\usepackage{mathtools}
\usepackage{multirow}

\usetikzlibrary{matrix}
\usetikzlibrary{calc}
\usetikzlibrary{shapes.geometric}
\usetikzlibrary{arrows.meta}
\usetikzlibrary{patterns}
\usetikzlibrary{patterns.meta}
\usetikzlibrary{backgrounds}

\begin{document}

\title{Removing Geometric Bias in One-Class Anomaly Detection with Adaptive Feature Perturbation}

\newcommand{\ands}{~~~~~~~~~}
\author{Romain Hermary
\ands Vincent Gaudillière
\ands Abd El Rahman Shabayek
\ands Djamila Aouada\\[.2cm]
University of Luxembourg, Esch-sur-Alzette, Luxembourg\\
{\tt\small \{romain.hermary, vincent.gaudilliere, abdelrahman.shabayek, djamila.aouada\}@uni.lu}
}
%
\maketitle
%
\begin{abstract}
    
One-class anomaly detection aims to detect objects that do not belong to a predefined normal class.
In practice, training data lack those anomalous samples; hence, state-of-the-art methods are trained to discriminate between normal and synthetically-generated pseudo-anomalous data.
Most methods use data augmentation techniques on normal images to simulate anomalies.
However, the best-performing ones implicitly leverage a geometric bias present in the benchmarking datasets.
This limits their usability in more general conditions.
Others are relying on basic noising schemes that may be suboptimal in capturing the underlying structure of normal data.
In addition, most still favour the image domain to generate pseudo-anomalies, training models end-to-end from only the normal class and overlooking richer representations of the information.
To overcome these limitations, we consider frozen, yet rich feature spaces given by pretrained models, and create pseudo-anomalous features with a novel adaptive linear feature perturbation technique.
It adapts the noise distribution to each sample, applies decaying linear perturbations to feature vectors, and further guides the classification process using a contrastive learning objective.
Experimental evaluation conducted on both standard and geometric bias-free datasets demonstrates the superiority of our approach with respect to comparable baselines.
The codebase is accessible via our \href{https://github.com/rhermary/PLUME}{public repository}\footnote{https://github.com/rhermary/PLUME}.
\end{abstract}
\section{Introduction}
\label{sec:intro}

Anomaly detection (AD), also known as novelty or out-of-distribution detection, is a widely investigated research topic, with applications ranging from machine faults detection~\cite{Serradilla2022DeepLM, MejriLRCGA24}, to malicious transactions in banking~\cite{Btoush2021ASO} and hazardous environmental situations in autonomous driving~\cite{Bogdoll2022AnomalyDI}.
In most cases, abnormal samples are too costly to obtain, and only normal samples are available.
This AD problem is therefore unsupervised and is also referred to as unlabelled one-class anomaly detection~\cite{CSI2020,UniConHA2023}, one-class novelty detection~\cite{SimpleNet2023}, or semantic outlier detection~\cite{CutPaste2021}.
In this paper, we tackle the problem from an image-level perspective, \textit{i.e.}, given images from a single semantic class, being able to classify unseen images as belonging to that class (\textit{normal} samples) or not (\textit{anomalies}).
This differs conceptually from anomaly segmentation, addressed in fields such as medical diagnosis~\cite{3464423} or industrial quality control~\cite{Bergmann2019MVTecA} and concerned with \textit{localisation} of pixel-level abnormalities within one semantic class, requiring dedicated heuristics and generally heavier computations.

\vspace{.12cm}
State-of-the-art (SotA) one-class anomaly detection approaches can be classified into three main categories: reconstruction-based methods~\cite{AnoGAN2017,NDA2021,OCGAN2019,ADGAN2019,DAGMM2018,TQM2019}, embedding-based methods~\cite{OCSVM2001,SVDD2004,Ruff2018DeepOC,DROCC2020,HRN2020} and synthetic anomaly-based methods~\cite{Geom2018,RotTrans2019,GOAD2020,CSI2020,iDECODe2022,SSD2021,DROC2021,UniConHA2023,PLAD2022}.
Reconstruction-based methods consist of training a deep neural network to reconstruct normal images and using the reconstruction error as a metric to differentiate between normal and abnormal samples.
However, such methods rely on the assumption that the network can accurately reconstruct training data but cannot generalise beyond it, which is not necessarily true in practice.
Embedding-based methods aim to embed normal feature distribution into a compressed space to facilitate the distinction with abnormal data.
However, these methods are subject to feature collapse and require dedicated heuristics to prevent this from happening.
\begin{figure}[t]
    \centering
    \vspace{-.05cm}
    \resizebox{\linewidth}{!}{\input{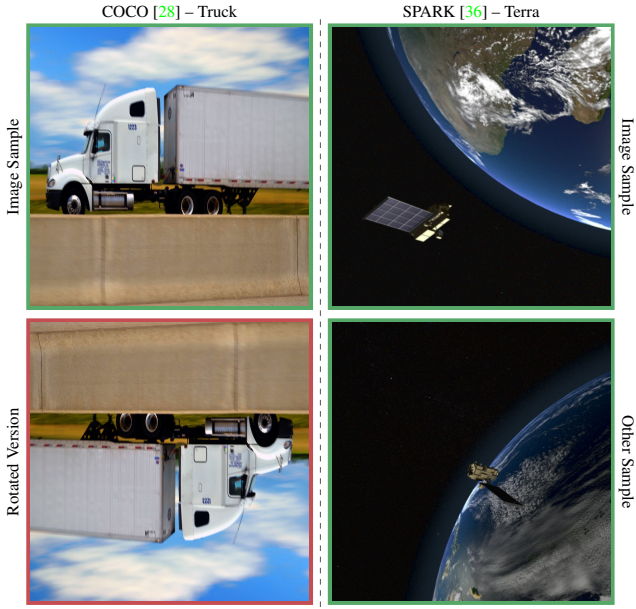}}
    \vspace{-.7cm}
    \caption{Illustration of the geometric bias. \textit{first row}: original images sampled from 2 public datasets.
    \textit{left column}: representative of CIFAR-10~\cite{CIFAR10} classes, second row is a manually rotated version of the top image and differs significantly from it.
    Rotated images can be treated as anomalies, since the original data is \textit{biased towards specific orientations}.
    \textit{right column}: second row contains another original image found in the dataset, which \textit{exhibits natural rotations} compared to the top image.
    In the absence of orientation bias, rotations cannot be used to create pseudo-anomalies.}
    \label{fig:showcase}
    \vspace{-.5cm}
\end{figure}

\vspace{-.22cm}
Synthetic anomaly-based methods consist mostly in generating artificial anomalies, \textit{i.e.} \textit{pseudo-anomalies}, from normal data during training to learn discriminative features able to separate normal and abnormal data.
These methods, whose development has been driven by the rise of self-supervised image representation learning~\cite{RotDet2018,Chen2020ASF}, have achieved among the highest performances on standard benchmarks such as CIFAR-10~\cite{CIFAR10}.
In practice, these methods generate pseudo-anomalous images using data augmentation techniques, and results from the literature show that the most effective type of augmentation is image rotation.
This is to the extent that, to the best of our knowledge, all of the best-performing methods take advantage of it~\cite{Geom2018,RotTrans2019,GOAD2020,CSI2020,iDECODe2022,SSD2021,DROC2021,UniConHA2023,Mirzaei_2024_CVPR}.
However, we believe that the effectiveness of such an image transform comes from the fact that original images in standard benchmarks are biased towards specific orientations (animals or cars are rarely pictured upside down).
This is supported by Goyal~\etal~\cite{DROCC2020} who noticed that these types of methods are \textit{``heavily domain dependent"}, and that even within the same dataset the \textit{``suitability of a transformation varies based on the structure of the typical points"}, making some transformations very powerful in some cases but ineffective in others.
In addition, Gidaris~\etal~\cite{RotDet2018} found that human-captured image are subject to \textit{well-posedness}: always being in the same orientation, making \textit{"the rotation recognition task well defined"}.
To the misfortune of SotA methods, this geometric bias is not omnipresent and therefore cannot be leveraged, for example, in the case of spacecraft datasets~\cite{SPARK}.
This intuition is illustrated in \Cref{fig:showcase}.

Instead of leveraging geometric bias in the images, SimpleNet~\cite{SimpleNet2023} creates pseudo-anomalies at the feature level; we argue that such an approach mitigates the risk of not being generalisable beyond standard anomaly detection datasets.
Nevertheless, this method is from the other branch of AD, localisation, it depends on locally applied perturbations that do not allow for relevant semantic anomalies to be generated, and, especially, relies on fixed Gaussian noise for every sample, which refrains from any adaptation to the likely complex structure of the representation space.
The latter limitation was anticipated with the method PLAD~\cite{PLAD2022}.
It proposes to automatically adapt the noise distribution parameters to every sample using a Variational Auto-Encoder (VAE)~\cite{VAE2014}.
However, such a method applies multiplicative and additive noise at the image level, and we argue that (i) deep learning features obtained from pretrained models represent a more optimal encoding of useful semantic information in images, making it easier to bound normality while also limiting the risks of overfitting to the normal class, and that (ii) applying a \emph{linear} noising process is better suited to perturb feature vectors.

Therefore, we propose \ours, \textit{Pertubation Learning with Unified inforMation Embeddings}, an unsupervised anomaly detection method that generates adaptive pseudo-anomalies within the feature space during training.
This is achieved by introducing a linear disturbance to the normal representations, automatically modulating this disturbance with an adaptive noise level.
Then, a multilayer perceptron is used to learn a decision boundary between normal and pseudo-anomalous features.
It is further guided by a contrastive learning objective to aggregate normal features in a more unified representation and repel abnormal ones.
In a nutshell, our contributions are three-fold:\\[-.65cm]
\begin{itemize}
    \item we propose to work in a frozen feature space and develop an adaptive linear feature perturbation technique to create pseudo-anomalies from normal samples without exploiting any dataset-specific geometric bias,\\[-.72cm]
    \item we propose \ours{}, a model that integrates the aforementioned feature perturbator to learn to detect anomalies in an unsupervised manner with the help of a contrastive learning objective,\\[-.72cm]
    \item we provide an experimental validation that demonstrates the superiority of our method with respect to comparable baselines on both standard and geometric bias-free datasets.\\[-.6cm]
\end{itemize}

The rest of the paper is organised as follows. Related work is discussed in \Cref{sec:related}, and our approach is described in \Cref{sec:method}. \Cref{sec:xp} presents experimental validation against state-of-the-art baselines, along with an ablation study. Specific aspects of our method are further discussed in \Cref{sec:disc}, and \Cref{sec:conclu} concludes the paper.
%
\section{Related Work}
\label{sec:related}

\subsection{Reconstruction \& Embedding-based Methods}

Reconstruction-based methods~\cite{AnoGAN2017,NDA2021,OCGAN2019,ADGAN2019,DAGMM2018,TQM2019} exploit the notion that anomalous image regions deviate significantly from the patterns observed in training data, making their faithful reconstruction challenging. 
These methods leverage generative models such as auto-encoders (AEs) or generative-adversarial networks (GANs).
They learn a compressed representation and reconstruct normal data from it. Deviations from this learnt representation are flagged as anomalies. However, learning to generate the entire normal data distribution from a finite training set can be challenging and inaccurate in practice (see \Cref{tab:comp}).

Embedding-based methods for AD embed features extracted from normal data into a lower-dimensional space. This compression allows for easier identification of anomalies, which supposedly lie far away from the cluster formed by normal features. 
Historically, OCSVM~\cite{OCSVM2001} uses kernel SVM to separate the normal data from the origin, considering it as the only negative data point.
Support Vector Data Description (SVDD)~\cite{SVDD2004} also uses kernel SVM to find an hypersphere that encloses the normal data.
However, such shallow methods struggle on complex domains like images, where feature-engineering is quite challenging.
Deep-SVDD~\cite{Ruff2018DeepOC} is a deep learning-based version of SVDD; it minimizes the volume of the hypersphere that encloses the normal features.
However, it may suffer from representation collapse.
HRN~\cite{HRN2020} tackles this issue by a holistic regularization method.
It constrains the model training to consider the normal features holistically.
Such a heuristic is, however, insufficient in practice (\Cref{tab:comp}).

\subsection{Synthetic Anomaly-based Methods}

Synthesising-based methods adopt a different approach, with the aim of learning how to differentiate normal data from synthesised anomalies. 

Several works~\cite{Geom2018,RotTrans2019,GOAD2020} aim to learn the observed geometric characteristic of the normal data (\textit{e.g.}, orientation of the object) by applying specific transformations (\textit{e.g.}, rotations) to the input image and learning to predict the parameters of the applied transformation.
The intuition is that, at test time, a failure to accurately predict the transform likely comes from geometric image properties that are different from those of normal data, \textit{i.e.} are from anomalies.
DROC~\cite{DROC2021} and CSI~\cite{CSI2020} leverage distributional augmentation (\textit{e.g.}, rotation) to simulate real-world outliers and model the inlier distribution by contrasting original samples with these simulated outliers.
SSD~\cite{SSD2021} also leverages contrastive learning.
However, the learnt representation is uniformly distributed on the hypersphere, contradicting the core principle of AD, which suggests that the inlier distribution should remain compact against outliers.
To avoid this, UniConHA~\cite{UniConHA2023} proposes a dedicated ``unilaterally aggregated'' contrastive learning objective.

Nevertheless, to be successful all these previous works critically rely on side-information in the form of appropriate transformations to generate pseudo-anomalies, which are dataset-dependent.
As the major SotA methods in one-class AD suggest, rotation transformations are relevant to create pseudo-anomalies on the standard benchmark.

In PLAD~\cite{PLAD2022}, pseudo-anomalies are generated by applying perturbations on the normal images, using multiplicative and additive noise whose parameters are learnt to be sample-specific.
A discriminator made of fully convolutional layers is then simultaneously trained to learn a tight boundary around normal data.
This generic method is therefore applicable to any dataset, since it does not rely on any specific bias such as uniform object orientation repeated across normal images.

However, in the era of foundation models~\cite{Sameni_2024_CVPR, Jia2021ScalingUV,Bordes2024AnIT}, where machines can perceive and understand the visual world with unprecedented accuracy, we argue that pre-trained features are more compact and structured representations of image content than raw pixel information and are therefore more suitable supports for perturbation-based anomaly synthesis.
Nevertheless, we demonstrate in what follows that additive and multiplicative noise, even adaptive, is not effective on deep features (\Cref{ssec:ablation}); therefore, we introduce with \ours{} a new adaptive linear perturabtion methodology that makes better use of this vector space, achieving new SotA results.
In addition, our method enables the reduction of the trained discriminator to a small multilayer perceptron, which can be seamlessly integrated into pre-existing architectures --sharing the feature extraction costs--, while maintaining strong AD performance through the use of a contrastive learning objective.

%
\section{Methodology}
\label{sec:method}
In this section, we detail our approach towards unsupervised semantic anomaly detection.
We consider a general data space $\ffield{X} \subset \ffield{R}^\fdim{D}$, where $\fdim{D}$ is any dimension, in which data points can either follow or defy an agreed normality. We denote by $\ffield{X}^+$ the subspace of normality and by $\ffield{X}^-=\ffield{X}\setminus\ffield{X}^+$ the subspace of anomalies. 
The training set $\fset{X}\subset\ffield{X}$ is partitioned into $\fset{X}^+\subset\ffield{X}^+$ (normal samples) and $\fset{X}^-\subset\ffield{X}^-$ (anomalies).
In this paper, we place ourselves in the extreme case where no anomalous sample is available during training, meaning our solution is designed for applications where $\fset{X}^- = \emptyset$ (unsupervised anomaly detection).
%
\subsection{Overview}

To detect divergent samples, we train a classifier to predict whether the sample comes from $\ffield{X}^+$ or $\ffield{X}^-$.
We focus on overcoming the lack of negative samples during the classifier training and improving existing fitting techniques.
More precisely, we leverage the use of a pre-trained feature extractor to work on an improved vectorised representation of the data, \textit{i.e.} \textit{features}, and adapt our method to be effective within the considered vector space.
In this work, we are therefore considering a model $\ffunc{h} : \ffield{R}^{\fdim{J}\times\fdim{H} \times\fdim{W}} \rightarrow \ffield{X}$ to go from the image space $\ffield{R}^{\fdim{J}\times\fdim{H} \times\fdim{W}}$ (with $\fdim{J}$, $\fdim{H}$, $\fdim{W}$ the number of channels, height and width of the images) to a feature space considered as our data space $\ffield{X}$.

The role of the classifier is to derive the complex decision boundary to establish the origin of the samples.
With only the positive class available during training, it would be unnecessary for the network to find a well-fitting boundary around $\fset{X^+}$ to satisfy the classification objective; instead, the modelled solution would ignore the data information, converging towards an infinitely large boundary radius and $100\%$ false negatives when encountering anomalies.

\begin{figure*}[t]
    \centering
    \vspace{-.3cm}
    \resizebox{.94\linewidth}{!}{\input{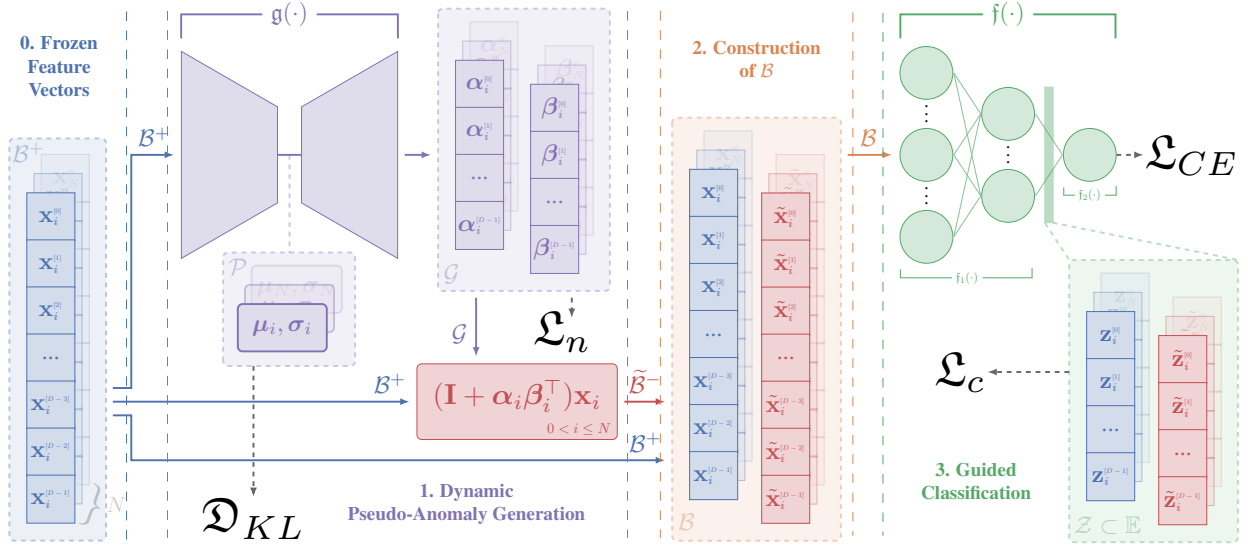}}\\[-.3cm]
    \caption{Diagram of our architecture, with the links between the different parts for a batch $\fset{B}^+$ of $\fdim{N}$ vectors of dimension $\fdim{D}$ (features extracted with the backbone $\ffunc{h}$). In purple, the perturbator $\ffunc{g}$ generating ($\fgrvm{\alpha}_i$, $\fgrvm{\beta}_i$) couples for each sample. In red, the application of linear perturbation to generate $\fset{\Tilde{B}}$ and construction of $\fset{B}$. In green, the classifier $\ffunc{f}$ separated in two parts to access the classifier embeddings, in $\ffield{E}$.}
    \label{fig:archi}
    \vspace{-.4cm}
\end{figure*}

To avoid this degenerate behaviour, we base our method on the union of $\fset{X}^+$ with generated negative samples.
To cope with the lack of anomalous samples in the training set $\ffield{X}^+$, we follow the standard paradigm that consists in applying a set of transformations $\fset{T}$ on elements of $\fset{X}^+$ to simulate anomalies.
The difficulty lies in determining what would be a valid $\fset{T}$ for any type of data.
Indeed, we have no guarantee that a transformation applied to a normal sample gives a legitimate anomaly, except if we have prior knowledge of both normal and abnormal data.

Following Cai and Fan~\cite{PLAD2022} (\plad{}), we propose to adaptively noise the normal data, with the rationale that there is a need for the noise to be adapted to each sample, since they all lie in a different positions in the data space, and that the noise level needs to be small for the perturbed samples to be valid and allow sane convergence.
Aiming for an effective generation process, we adopt their adversarial strategy with two models trained jointly.

\Cref{fig:archi} depicts an overview of our architecture.
A first model, called the \textit{perturbator} (purple), generates the parameters for the transformations applied to the normal samples.
Eventually, its objective is to model the parameter distribution which will allow correct transformations to be applied.
In \ours{}, we design $\fset{T}$ to be effective in a feature space.

The second model is the classifier (green), for which the training objective is to distinguish between normal samples and pseudo-anomalies.
Inspired by contrastive learning strategies, we make the model more efficient in separating samples from different sources.
By forcing the classifier to project normal data into a unified representation, we aim at easing the task of deriving the decision boundary.

\Cref{ssec:featpert} further describes the perturbator module and \Cref{ssec:disc} the optimisations in place for the classifier.

\subsection{Feature Perturbator}
\label{ssec:featpert}

Inspired by \plad{}, a set of pseudo-anomalies $\Tilde{\fset{X}}$ is dynamically generated from $\fset{X}^+$ using a trainable neural network $\ffunc{g} : \ffield{R}^\fdim{D} \rightarrow \ffield{R}^\fdim{D}\times\ffield{R}^\fdim{D}$, the perturbator.
In \plad{}, the perturbator generates a couple of vectors $(\fgrvm{\alpha}_i, \fgrvm{\beta}_i) = \ffunc{g}(\fvm{x}_i)$ specific to each normal sample $\fvm{x}_i \in \fset{X}^+$ (vectorised image) and interpreted as element-wise multiplicative and additive noise maps ($\fgrvm{\alpha}_i \odot \fvm{x}_i + \fgrvm{\beta}_i$).
These elements are restrained not to produce pseudo-anomalies lying far away from normal samples --implied by the adversarial objective of the classifier--, by a simple yet effective constraint $\ffunc{L}_n^{(i)}$, defined for the $i^{\textrm{th}}$ sample in the current batch $\fset{B}^+$ as:
\begin{equation}
\label{eq:noise}
    \ffunc{L}_n^{(i)} = \lVert \fgrvm{\alpha}_i - \fvm{1}\rVert^2 + \lVert \fgrvm{\beta}_i - \fvm{0} \rVert^2,
\end{equation}
where $\mathbf{1} = [1, 1,\dots, 1]^\top$ and $\mathbf{0} = [0, 0,\dots, 0]^\top$ are vectors of dimension $\fdim{D}$, same as $\fvm{x}_i$, $\fgrvm{\alpha}_i$ and $\fgrvm{\beta}_i$. $\lVert \cdot \rVert$ is the $\ell^2$ norm.

Although shown to perform properly on images, additive and multiplicative noises are mostly used on raw signal data and we argue that such transformations are not optimal in a feature vector space.
To make a more sensible use of the features produced by $\ffunc{h}$, better preserve the geometry, structure and integrity of the data space while enabling more complex transformations, we propose to redefine the perturbation as a linear map $\fvm{A}_i \in \ffield{R}^{\fdim{D}\times\fdim{D}}$.
For the perturbation to produce cohesive and non aberrant data, we want this map to represent a small perturbation, hence be close to the identity map.
Instead of modifying the perturbator $\ffunc{g}$ to generate directly a matrix in $\ffield{R}^{\fdim{D}\times\fdim{D}}$, which would heavily impact its number of parameters and disturb its convergence, we propose to keep $\fgrvm{\alpha}_i$, $\fgrvm{\beta}_i$ (and their constraint $\ffunc{L}_n^{(i)}$), using them ingenuously to instead noise an identity matrix $\fvm{I} \in \ffield{R}^{\fdim{D}\times\fdim{D}}$:
\begin{equation}
    \fvm{A}_i = \fvm{I} + \fgrvm{\alpha}_i \fgrvm{\beta}_i^\top.
\end{equation}
During the training, the perturbation $\fvm{A}_i$ then converges towards the desired identity map $\fvm{I}$, and pseudo-anomalies can be generated by applying linear perturbations:
\begin{equation}
    \Tilde{\fvm{x}}_i = \fvm{A}_i \fvm{x}_i.
\end{equation}

The batch $\fset{B}$ given to the classifier in any training step is therefore the union of a subset $\fset{B}^+=\left\{\fvm{x}_i\right\}_{i=1}^N \subset \fset{X}^+$ of $\fdim{N}$ normal samples and the generated set of pseudo-anomalies $\widetilde{\fset{B}}^- = \left\{\Tilde{\fvm{x}}_i\right\}_{i=1}^N$.

To generate $\fgrvm{\alpha}_i$ and $\fgrvm{\beta}_i$, we use a Variational Auto-Encoder (VAE)~\cite{VAE2014} since shown to perform well as the generator $\ffunc{g}$~\cite{PLAD2022}.
It is decomposed into two parts: an encoder and a decoder.
The first aims to fit a noise distribution, generating the set of intermediary parameters $\fset{P} = \left\{(\fgrvm{\mu}_i, \fgrvm{\sigma}_i)\right\}_{i=1}^\fdim{N}$, with $\fgrvm{\mu}_i$ and $\fgrvm{\sigma}_i\in \ffield{R}^\fdim{D}$ the encoded mean and variance for the $i^\textrm{th}$ sample.
The second generates the set of transformation parameters $\fset{G} = \left\{(\fgrvm{\alpha}_i, \fgrvm{\beta}_i)\right\}_{i=1}^\fdim{N}$, also respective to each sample.
As in the original work of Kingma and Welling~\cite{VAE2014}, the encoder is primarly optimised with a Kullback-Leibler divergence $\ffunc{D}_{KL}^{(i)}$:
\begin{equation}
    \label{eq:dkl}
    \ffunc{D}_{KL}^{(i)} = \frac{1}{2}\sum\limits_{d=1}^\fdim{D} \left[\fgrvm{\sigma}_i^2 + \fgrvm{\mu}_i^2 - 1 - \mathrm{log}(\fgrvm{\sigma}_i^2)\right]_d,
\end{equation}
where $[\cdot]_d$ denotes the value at the $d^\textrm{th}$ element of the vector.
The objective of the decoder parameters is modified and the latter are optimised owing to $\mathfrak{L}_n^{(i)}$ (Eq.~\ref{eq:noise}).
\subsection{Classifier}
\label{ssec:disc}

We use a neural network $\ffunc{f} : \ffield{R}^\fdim{D} \rightarrow \ffield{R}$ to distinguish between normal data and anomalies.
As a training objective, we define a target value $y \in \{0, 1\}$ equal to $1$ for all $\fvm{x} \in \fset{X}^+$ and $0$ for all generated $\Tilde{\fvm{x}}$ passed to $\ffunc{f}$.
Subject to a logistic activation function, any prediction $\hat{y}$ of the classifier lies in the range $[0,1]$ and a binary cross-entropy loss $\ffunc{L}_{CE}^{(i)}$ is used as a training objective. It is defined as:
\begin{equation}
    \ffunc{L}_{CE}^{(i)} = -\left[y_i\mathrm{log}(\hat{y}_i) + (1 - y_i)\mathrm{log}(1 - \hat{y}_i)\right],
\end{equation}
for the $i^\textrm{th}$ sample in the current batch $\fset{B}$, and with $\hat{y}_i$ and $y_i$ the predicted and groundtruth sample values, respectively.

\noindent To facilitate the derivation of the decision boundary,
our intuition is that the classifier $\ffunc{f}$ should map the normal data to a very dense space, \textit{far} from where anomalies would be projected.

To develop the idea, let us decompose $\ffunc{f}$ into $\ffunc{f}_1$ and $\ffunc{f}_2$ so that $\ffunc{f} = \ffunc{f}_2 \circ \ffunc{f}_1$.
In our neural network setup, $\ffunc{f}_1 : \ffield{R}^\fdim{D} \rightarrow \ffield{E}$ represents the first layers of the network that condense the information from the input into an embedding vector $\fvm{z} \in \ffield{E}$.
Ideally, all vectors from the classifier embedding space $\ffield{E}$ should be very informative about the origin of the data.
On the other hand, we have $\ffunc{f}_2 : \ffield{E} \rightarrow \ffield{R}$ representing the last layer that combines all the information extracted into the final decision and is thus the boundary estimator.
The more $\ffunc{f}_1$ is discriminative, the easier it is for $\ffunc{f}_2$ to model the decision boundary.

In order to increase the separability of the embeddings in $\ffield{E}$, depending on their origin, we propose to maximise the similarity of every $\fvm{z}$ extracted from a normal sample while, at the same time, maximise their dissimilarity with the other embeddings extracted from anomalous samples.
As a measure of similarity between the vectors, we follow the literature~\cite{Hinton2015DistillingTK, Chen2020ASF} and leverage the use of temperature-scaled cosine similarity $\ffunc{s} : \ffield{E} \times \ffield{E} \rightarrow [-\frac{1}{\tau}, \frac{1}{\tau}]$:
\begin{equation}
\label{eq:cossim}
    \ffunc{s}(\fvm{v}_1, \fvm{v}_2) = \frac{1}{\tau} \frac{\fvm{v}_1 \cdot \fvm{v}_2}{\lVert \fvm{v}_1 \rVert \lVert \fvm{v}_2 \rVert},
\end{equation}
where $\tau \in \ffield{R}$ helps controlling the concentration level of the similarities distribution~\cite{Wu2018UnsupervisedFL}. To simultaneously optimise both objectives, we make use of the contrastive loss defined in \Cref{eq:contrastive}, where $\fvm{z}_i$ denotes the embedding vectors derived from normal samples, $\Tilde{\fvm{z}}_i$ the ones from the generated pseudo-anomalies, and $\mathbbm{1}_{a\neq b}$ is an indicator function evaluating to $1$ if $a\neq b$, $0$ otherwise:
\begin{multline}
\label{eq:contrastive}
\ffunc{L}_{c}^{(i)} = \frac{1}{\fdim{N} - 1}\sum\limits_{j=1}^\fdim{N}-\mathbbm{1}_{j\neq i}\mathrm{log}
\Biggr[\\
\frac{\mathrm{exp}(\ffunc{s}(\fvm{z}_i, \fvm{z}_j))}{\sum_{l=1}^{\fdim{N}}\left[\mathrm{exp}(\ffunc{s}(\fvm{z}_i, \Tilde{\fvm{z}}_l)) + \mathbbm{1}_{l\neq i}\mathrm{exp}(\ffunc{s}(\fvm{z}_i, \fvm{z}_l))\right]}\Biggr].
\end{multline}

Finally, our loss $\ffunc{L}$, defined in \Cref{eq:totloss}, is an average over the batch of the summed sub-losses. $\lambda, \nu$ and $\gamma \in \ffield{R}$ are tunable hyperparameters used to balance the different losses with respect to $\ffunc{L}_{CE}$:
\begin{equation}
\label{eq:totloss}
    \ffunc{L} = \frac{1}{\fdim{N}}\sum\limits_{i = 1}^\fdim{N} \left(\ffunc{L}_{CE}^{(i)} + \lambda\ffunc{L}_n^{(i)} + \nu\ffunc{D}_{KL}^{(i)} + \gamma\ffunc{L}_c^{(i)}\right).
\end{equation}
%
\section{Experimental Evaluation}
\label{sec:xp}

In this section, we delve into the evaluation protocol used and present the results obtained from our experiments. We compare our approach with SotA methods and provide additional experiments to highlight its added value.

\subsection{Datasets and Evaluation}

Following the main benchmark in unsupervised one-class AD~\cite{HRN2020, PLAD2022, Ruff2018DeepOC}, we evaluate our method on CIFAR-10~\cite{CIFAR10} and report the best AUCs.
We produce and give additional results on CIFAR-100~\cite{CIFAR10} and SPARK~\cite{SPARK}, the latter being used as a non-geometrically-biased and application-oriented dataset.
To the best of our knowledge, our work is the first to exploit SPARK in the context of AD.\\

\begin{figure}[t]
    \centering
    \resizebox{.9\linewidth}{!}{\input{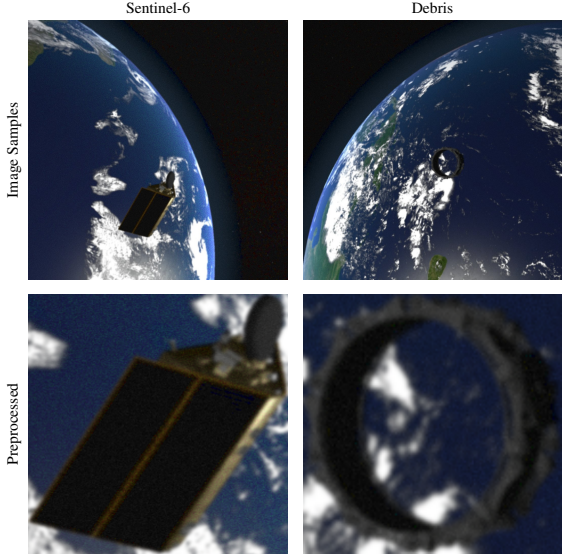}}
    \vspace{-.2cm}
    \caption{SPARK dataset~\cite{SPARK} samples: satellite (left) and debris (right); original (top) and cropped images (bottom).}
    \label{fig:sparksamples}
    \vspace{-.4cm}
\end{figure}

\noindent\textbf{CIFAR-10 \& CIFAR-100~\textnormal{\cite{CIFAR10}}.}
CIFAR-10 dataset is composed of $10$ object classes. For each of them, there are $5,000$ training samples. The validation set is composed of $10,000$ samples balanced between all classes. Size of images is 32$\times$32. We evaluate our method following the one-class classification paradigm, \textit{i.e.} each class is alternately considered as the normal class, while the 9 remaining classes form the set of anomalies. CIFAR-100, on the other hand, has $100$ object classes grouped into $20$ meta-classes. Each meta-class has $2500$ samples of $5$ different objects. The test set is also made up of $10000$ images, and we alternatively consider each meta-class as a normal class and the remaining ones as anomalies.\\[-.26cm]

\noindent\textbf{SPARK~\textnormal{\cite{SPARK}}.}
SPARK is a synthetic dataset composed of images of satellites and debris orbiting the Earth.
The $10$ satellite classes ($75,000$ training images) are combined as one normal class, while the debris class (5 different debris models) is used during validation as anomaly class.
The validation dataset is then composed of $5,000$ debris images and $25,000$ satellite images.
We only use the $\fdim{RGB}$ images and discard depth maps.
As we focus on the semantics of the objects, we also crop the images to the ground-truth bounding boxes and reshape the resulting images into a square format ($512\times512$) using bilinear interpolation.
Sample images are provided in \Cref{fig:sparksamples}.\\

\begin{figure}[t]
    \centering
    \includegraphics[width=\linewidth]{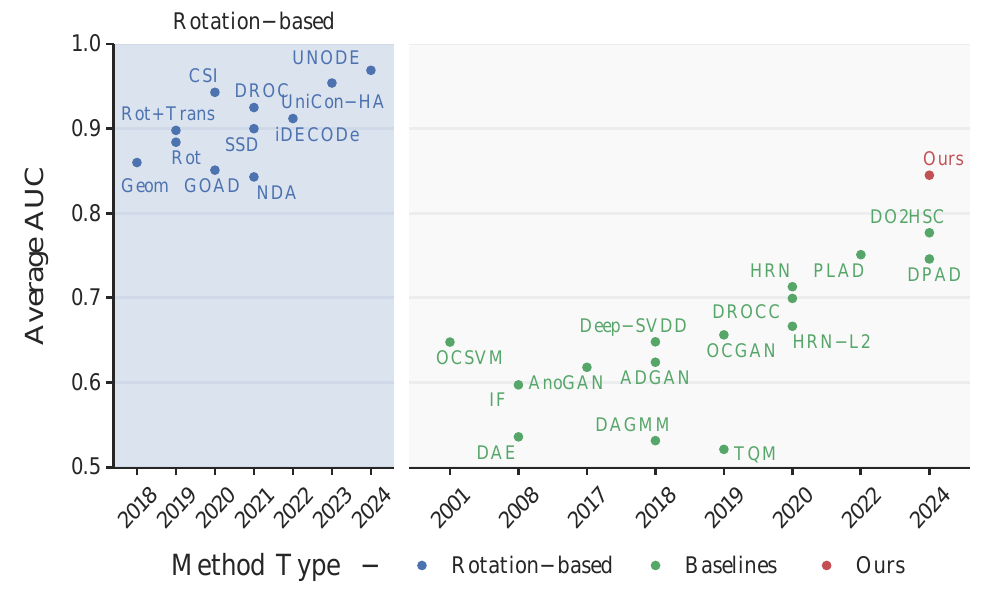}
    \vspace{-.7cm}
    \caption{Performance of SotA methods on CIFAR-10 dataset. Methods on the left (blue background) achieve among the best results by exploiting a geometric bias in the dataset, \textit{i.e.} typical object orientations. On the contrary, methods on the right (grey background - green dots) are less dataset-specific and are therefore considered as our baselines. Detailed results of all methods can be found in the supplementary.
    }
    \label{fig:baselines}
    \vspace{-.5cm}
\end{figure}
%
\noindent\textbf{Metrics.} To avoid relying on a specific threshold, the standard evaluation procedure consists in reporting the best Area Under ROC-Curve (AUC)~\cite{Bradley1997TheUO} during the validation stage. We thus use this metric in our experiments.

\subsection{Baselines}

We compare \ours{} on CIFAR-10 with 9 baseline methods.
To the best of our knowledge, this selection covers most SotA approaches and contains the most accurate ones.
We do not consider rotation-based methods, that exploit a bias present in the dataset but not necessarily in real-life scenarios.
However, for the sake of transparency and completeness, we have depicted all the results in \Cref{fig:baselines}, also including older well-known methods.
We also ran \ours{} and our principal baseline PLAD~\cite{PLAD2022} on CIFAR-100 and SPARK, and reported the results in \Cref{tab:cifar100}.

\begin{table*}
\begin{subtable}[t]{\linewidth}
  \centering
  \resizebox{\linewidth}{!}{
    
\begin{tabular}{l|cccccccccc|c}
        \hline &&&&&&&&&& \\[-1em]
        Method &  Plane & Car & Bird & Cat & Deer & Dog & Frog & Horse & Ship & Truck & Mean \\
        \hline\hline &&&&&&&&&& \\[-1em]
        ADGAN~\cite{ADGAN2019}    &  63.2 & 52.9 & 58.0 & 60.6 & 60.7 & 65.9 & 61.1 & 63.0 & 74.4 & 64.2 & 62.4\\
        OCGAN~\cite{OCGAN2019}   &   75.7 & 53.1 & 64.0 & 62.0 & 72.3 & 62.0 & 72.3 & 57.5 & 82.0 & 55.4 & 65.6 \\
        TQM~\cite{TQM2019} &  40.7 & 53.1 & 41.7 & 58.2 & 39.2 & 62.6 & 55.1 & 63.1 & 48.6 & 58.7 & 52.1\\
        DROCC~\cite{DROCC2020} &  79.2 & 74.9 & 68.3 & 62.3 & 70.3 & 66.1 & 68.1 & 71.3 & 62.3 & 76.6 & 69.9\\
        HRN-L2~\cite{HRN2020}    &  80.6 & 48.2 & 64.9 & 57.4 & 73.3 & 61.0 & 74.1 & 55.5 & 79.9 & 71.6 & 66.7\\
        HRN~\cite{HRN2020}      &   77.3 & 69.9 & 60.6 & 64.4 & 71.5 & 67.4 & 
        77.4 & 64.9 & 82.5 & 77.3 & 71.3\\
        DPAD~\cite{Fu2024DensePF} & 78.0\stdv{0.3} & 75.0\stdv{0.2} & 68.1\stdv{0.5} & 66.7\stdv{0.4} & 77.9\stdv{0.8} & 68.6\stdv{0.3} & 81.2\stdv{0.4} &  74.8\stdv{0.2} & 79.1\stdv{1.0} & 76.1\stdv{0.2} & 74.6\\
        DO2HSC~\cite{Zhang2023DeepOH} & 81.3\stdv{0.2} & 82.7\stdv{0.3} & 71.3\stdv{0.4} & 71.2\stdv{1.3} & 72.9\stdv{2.1} & 72.8\stdv{0.2} & 83.0\stdv{0.6} & 75.5\stdv{0.4} & 84.4\stdv{0.5} & 82.0\stdv{0.9} & 77.7\\
        PLAD~\cite{PLAD2022} & 82.5\stdv{0.4} & 80.8\stdv{0.9} & 68.8\stdv{1.2} & 65.2\stdv{1.2} & 71.6\stdv{1.1} & 71.2\stdv{1.6} & 76.4\stdv{1.9} & 73.5\stdv{1.0} & 80.6\stdv{1.8} & 80.5\stdv{0.3} & 75.1 \\
        \hline\hline &&&&&&&&&& \\[-1em]
  \textbf{\ours{}} & \textbf{89.5}\stdv{1.0} & \textbf{85.7}\stdv{1.7} & \textbf{74.5}\stdv{2.3} &  \textbf{78.3}\stdv{3.6} & \textbf{87.7}\stdv{1.3} &  \textbf{79.5}\stdv{3.7} & \textbf{87.5}\stdv{0.9} & \textbf{84.6}\stdv{3.7} & \textbf{87.8}\stdv{3.1} & \textbf{90.0}\stdv{2.7} & \textbf{84.5}\\
  \textbf{\ours{}}~~~~(Max.) & 91.0 & 90.1 & 80.2 & 80.8 & 90.6 & 83.7 & 88.4 & 89.5 & 92.6 & 94.0 & 88.1\\
    \hline
    \end{tabular}

  }
  \caption{Anomaly detection results on CIFAR-10. We report average results over $5$ runs for each class, with the best results per class in bold. Maximum AUC values for each class are reported in the last row (not considered in the comparison). The best performing value of $\lambda$ was selected for each class.\\[-.2cm]}
    \label{tab:comp}
\end{subtable}
\begin{subtable}[t]{0.2\linewidth}
  \vspace{-1cm}
  \resizebox{\linewidth}{!}{
  \centering
    
\begin{tabular}{c|c}
\hline
 Method & Average AUC\\
\hline\hline
 PLAD\textsuperscript{$\dagger$} & 63.8\\
 \textbf{\ours{}} & \textbf{80.3}\\
\hline
\end{tabular}

  }
  \caption{Average AD score on CIFAR-100. We report average results over the 20 meta-classes and 5 runs ($\lambda = 5$). We report our results ($\dagger$) with the available implementation of PLAD~\cite{PLAD2022}.}
    \label{tab:cifar100}
\end{subtable}
\hspace{.1cm}
\begin{subtable}[t]{0.78\linewidth}
    \resizebox{\linewidth}{!}{
        
\begin{tabular}{c|c|cccccccccc|c}
  \hline
        Perturbation & Guidance & Plane & Car & Bird & Cat & Deer & Dog & Frog & Horse & Ship & Truck & Mean \\
  \hline\hline &&&&&&&&&&&&\\[-1em]
Gaussian & -  & 83.3\stdv{2.6} & 75.7\stdv{10.0} & 69.8\stdv{3.2} & 72.8\stdv{9.1} & 80.6\stdv{3.4} & 72.0\stdv{3.3} & 84.5\stdv{1.7} & 79.6\stdv{5.7} & 79.9\stdv{3.6} & 76.9\stdv{9.3} & 77.5\\

Gaussian & $\checkmark$ & 75.1\stdv{3.2} & 79.6\stdv{3.4} & 68.4\stdv{1.3} & 70.4\stdv{1.6} & 74.5\stdv{3.7} & 69.1\stdv{2.6} & 78.5\stdv{2.1} & 69.1\stdv{3.2} & 75.2\stdv{5.0} & 77.6\stdv{5.3} & 73.7 \\
\hline
AddMult & - & 66.8\stdv{3.5} & 70.5\stdv{9.5} & 67.4\stdv{1.5} & 64.9\stdv{3.6} & 67.5\stdv{5.4} & 67.0\stdv{3.5} & 70.6\stdv{7.8} & 68.1\stdv{7.7} & 72.5\stdv{4.6} & 68.7\stdv{5.2} & 68.4\\ 
AddMult & $\checkmark$ & 59.2\stdv{2.6} & 62.6\stdv{2.6} & 61.2\stdv{1.5} & 58.7\stdv{2.0} & 63.0\stdv{1.9} & 62.6\stdv{1.0} & 68.6\stdv{3.0} & 59.5\stdv{4.9} & 61.1\stdv{1.2} & 57.5\stdv{4.6} & 61.4\\ 

\hline

LinearMap & - & 69.5\stdv{9.6} & 75.7\stdv{10.1} & 76.2\stdv{1.9} & 58.8\stdv{8.7} & 74.5\stdv{6.7} & 78.4\stdv{2.0} & 85.1\stdv{1.0} & 78.0\stdv{5.4} & 85.6\stdv{3.4} & 86.5\stdv{2.5} & 76.8 \\ 
LinearMap & $\checkmark$ (Mean) & 81.4\stdv{6.0} & 72.4\stdv{7.2} & 74.8\stdv{3.5} & 60.3\stdv{7.9} & 74.5\stdv{5.5} & 77.9\stdv{1.8} & 86.4\stdv{1.8} & 82.6\stdv{2.7} & 84.0\stdv{3.1} & 74.3\stdv{11.9} & 76.9\\

LinearMap & $\checkmark$ & 86.8\stdv{1.3} & 85.7\stdv{1.7} & 74.5\stdv{2.3} & 75.0\stdv{6.8} & 87.4\stdv{1.7} & 77.0\stdv{2.7} & 86.1\stdv{1.7} & 84.0\stdv{2.6} & 87.8\stdv{3.1} & 89.6\stdv{3.2} & 83.4\\ 

   \hline
\end{tabular}

    }
    \caption{Ablation study over the perturbation method and the contrastive guidance. Experiments on CIFAR-10 ($\ffunc{h}$: ResNet50, $\lambda = 5$, $5$ runs). We report results with adaptive additive and multiplicative noise (AddMult), adaptive linear mapping (LinearMap) and random Gaussian noise (Gaussian); with ($\checkmark$) and without (-) the contrastive loss, and results with the contrastive loss with respect to the mean embedding vectors of normal and pseudo-anomalous data (Mean).}
    \label{tab:ablation}
\end{subtable}
\vspace{-.3cm}
\caption{Results on (CIFAR-10 \Cref{tab:comp}), CIFAR-100 (\Cref{tab:cifar100}) and ablation study (\Cref{tab:ablation}).\\[-.3cm]}
\end{table*}

\subsection{Implementation Details}

We use the Cyclical Learning Rate (CLR)~\cite{Smith2015CyclicalLR, Smith2018SuperconvergenceVF} for optimal training phases and setting initial learning rate, in combination with the AdamW~\cite{Loshchilov2017DecoupledWD} optimiser.

As our feature extractor $\ffunc{h}$ and unless otherwise specified, we use the widely adopted ResNet50~\cite{ResNet} model pre-trained on ImageNet~\cite{ImageNet}.
This backbone is used as reference, but can be replaced by any other state-of-the-art model.
We emphasise the fact that we do not retrain $\ffunc{h}$ for any of the experiments. The other elements are trained from scratch.
An adaptive pooling layer is then used to bring every feature vector to a size of $\fdim{D} = 3072$, corresponding to the input size of the perturbator.
Subsequently, a batch normalisation layer is used to regularise the input space.
The batch size is set to $\fdim{N}=32$.
For the classifier, we use a simple multi-layer perceptron of 3 layers.
It is important noting that the perturbator is only used during training, and discarded when evaluating the model.
More details on the implementation can be found in the supplementary material.

We run each experiment $5$ times for $100$ epochs and, following the standard protocol in one-class AD~\cite{HRN2020, DROCC2020, PLAD2022, SimpleNet2023} which avoids epoch fine-tuning alongside benchmarking, save the best-performing model throughout the training.
Following PLAD, we set $\nu=1$ and tested different values for $\lambda$ (5, 10 and 20 in our experiments), and found that \ours{} is very stable and achieves SotA results whatever the value.
Detailed performance for each $\lambda$ can be found in the supplementary.

\subsection{Anomaly Detection on CIFAR-10 \& 100}

\Cref{tab:comp} presents a comprehensive evaluation of our proposed AD method \ours{}, along with comparable existing approaches on the CIFAR-10 dataset.
We first notice that \ours{} consistently outperforms all methods in terms of mean AUC, showcasing its superior AD capabilities.
For example, while OCGAN achieves a mean AUC of 65.6\%, DROCC 69.9\%, DO2HSC 77.7\% and PLAD 75.1\%, \ours{} significantly surpasses them with a mean AUC of 84.5\%.
Furthermore, our method also demonstrates consistent superiority in individual categories, achieving highest AUC on each class.

Additionally, the results on CIFAR-100 (\Cref{tab:cifar100}) expose the higher scalability of \ours{} to more heterogeneous normal classes and fewer training samples, without modification or additional training cost, compared to PLAD.
The detailed results can be found in the supplementary material.
These results underscore the substantial advancement in AD performance offered by \ours{}

\subsection{Ablation Study}
\label{ssec:ablation}

We report the impact of \ours{} components 
in \Cref{tab:ablation}.
ResNet50 features were used and hyperparameters are the same for the entire study ($\lambda = 5$).

We first replaced the linear mapping of \ours{} by random Gaussian noise (Gaussian).
Without contrastive loss, the model achieves an average AUC comparable to the current best unbiased baseline, DO2HSC~\cite{Zhang2023DeepOH}.
These results emphasise on the importance of better representations, here given by a ResNet50 trained on ImageNet.
However, the decrease in performance with the addition of contrastive guidance (almost 4pp of average AUC) reinforces the idea of unfitting pseudo-anomaly generation, apparently conflicting with the structure of the vector space.
Although the use of perturbations in the feature space seems very efficient, the simplicity and randomness do not allow to create sensible anomalous vectors that can be efficiently contrasted with respect to normal vectors.

Adaptive perturbations would allow to create meaningful pseudo-anomalous vectors, which in turn would profit the classifier and derive a better understanding of the normal data, and a better decision boundary.
However, basic additive and multiplicative noise (AddMult) performs poorly on vector representations of the data and only achieves 68.4\% average AUC.
In addition, it also shows a decrease in performance with the contrastive guidance: AddMult also seems not enough efficient at preserving strcuture coherence in the vector space and rather confuses the guided network.
Finally, our proposed linear mapping (LinearMap) is used and also achieves a performance comparable to PLAD~\cite{PLAD2022} and DO2HSC~\cite{Zhang2023DeepOH}.
The major difference from the other tested perturbation methods is that, thanks to its greater expressiveness and richer transformation set, adding contrastive guidance in this case is highly beneficial, offering an increase of 6.6pp of average AUC and reaching SotA results even without tuning $\lambda$.
While using a simpler version of the contrastive loss (Mean), \textit{i.e.} contrasting with respect to the mean embedding vectors of normal and pseudo-anomalous data, could be more computationally efficient (equation provided in the supplementary), it does not bring enough additional knowledge and guidance to the classifier.
A completely contrastive guidance (Eq.~\ref{eq:contrastive}) is necessary.

Overall, the ablation study indicates that (i) the AddMult perturbation performs worse  than simply adding non-adaptive Gaussian noise in a frozen feature space and is inferior to the LinearMap; (ii) the latter greatly profits from the contrastive guidance and both combined allow \ours{} to be successful in deriving a more precise boundary in the feature space, achieving SotA performance; (iii) using better representations, \ours{} determines this decision boundary by only training a 3-layers MLP.
These results underscore the effectiveness of the proposed methodology.
Further qualitative analysis on this structural difference can be found in the supplementary.

\subsection{Space Debris Detection on SPARK}
\begin{table}[t]
    \centering
    \resizebox{\linewidth}{!}{
        
\begin{tabular}{c|c|c|c|c}
    \hline
    & PLAD\textsuperscript{$\dagger$} & \multicolumn{3}{c}{\ours{}}\\
    \hline\hline
    Classifier & LeNet5~\cite{LeCun1998GradientbasedLA} & \multicolumn{3}{c}{3 Fully-Connected Layers}\\
    \hline
    \multirow{2}{*}{Input Data} & \multirow{2}{*}{Raw Images} & \multicolumn{3}{c}{Features} \\
    \cline{3-5}
    && ResNet50~\cite{ResNet} & VGG16~\cite{VGG16} & ConvNeXt~\cite{ConvNeXt}\\
  \hline\hline
   AUC & 0.69 & 0.77 & 0.74 & 0.76 \\
   \hline
\end{tabular}

    }\\[-.2cm]
    \caption{Anomaly detection results on SPARK. We report the best AUC reached by PLAD, which we ran ($\dagger$), and \ours{} with different frozen backbone architectures. We set $\lambda = 5$. ResNet50, VGG16 and ConvNeXt networks were pretrained on ImageNet.\\[-1cm]}
    \label{tab:spark_table}
\end{table}

SPARK's complexity is significantly higher compared to CIFAR-10, due to the similarities between normal and abnormal samples (space debris are often parts of spacecrafts) and to the common background information shared by all classes. Moreoever, it is important noting that 10 different spacecraft classes are combined in the normal class, adding an additional layer of intricacy to the data. This makes the unsupervised AD task particularly challenging.

The results in \Cref{tab:spark_table} illustrate that, despite this complexity, our method demonstrates a notable level of accuracy.
This is particularly impressive given that our classifier head is relatively simple and can easily fail to derive a proper decision boundary.
Moreover, none of the backbones was fine-tuned on datasets similar to SPARK (only pretrained on ImageNet). Therefore, their ability to extract optimal information from SPARK images is not guaranteed.

We trained PLAD, the main baseline, and obtained an AUC of 69\%.
Comparatively, \ours{} performs consistently better with the different feature extractors, reaching a maximum AUC of 77\% with ResNet50.
The accuracy achieved by \ours{} underlines the robustness of our method in dealing with complex datasets and the potential it holds for practical applications.

%
\section{Discussion \& Future Work}
\label{sec:disc}

PLUME has shown versatility and robustness, performing well with standard pre-trained backbones and complex data.
The ability to select different backbone architectures enables customisation for specific applications, while its universal design supports its scalability across different domains.
The key strengths of our method are its efficiency, seamlessly integrating with existing backbones already used for other tasks like pose or trajectory estimation, and its independence \textit{w.r.t.} data modality or special attributes.
This is especially valuable in resource-limited environments, since adding a small classifier for AD significantly optimises resource usage.

While the introduction of linear mapping increases memory costs during training, this is mitigated when pre-extracting features, as PLUME operates on a frozen feature space, also inferring a reduced processing time.
The computational overhead at test time is also minimal, since the AD network is a small MLP.
However, pre-trained backbones on specialised datasets may be needed for optimal performance in specific cases.

Future improvements could focus on refining our loss function, as the current noise constraint was adapted from previous work and not tailored to our perturbation method.
Maximising pairwise similarity between normal and generated samples, instead of constraining noise, could boost performance. Additionally, reducing the current four learning objectives may improve network stability.
Investigating extensions to time-series, facing time dependencies, offers another promising direction for future research.
%
\section{Conclusion}
\label{sec:conclu}
We have presented \ours, a novel approach for one-class anomaly detection.
It focuses on generating pseudo-anomalies within the feature space, leveraging a novel adaptive linear feature perturbation technique, without taking advantage of the geometric biases present in some datasets.
To guide the classification process, we introduced a contrastive learning objective that enhances the aggregation of normal features and repels abnormal ones.
This combination of adaptive feature perturbation and contrastive learning objective contributes to the effectiveness of our approach in distinguishing between normal and abnormal samples.
Our extensive experimental evaluation demonstrates the superiority of \ours{} over comparable baselines on both standard and geometric bias-free datasets.
By outperforming existing methods on both, our approach showcases its potential for real-world applications.
We believe it has the potential to contribute significantly to various domains, providing valuable insights and practical solutions for AD challenges.

\vspace{.1cm}
\noindent \textbf{Acknowledgments.} This research was funded in whole or in part by the Luxembourg National Research Fund (FNR), grant reference DEFENCE22/17813724/AUREA.

{\small
\bibliographystyle{ieee_fullname}
\bibliography{bib}
}
\clearpage
\appendix

\maketitle
\section{Experimental Environment}

Experiments were conducted with Python $3.10$~\cite{Rossum2009Python3R}, PyTorch $1.12.1$~\cite{Paszke2019PyTorchAI} (compiled for hardware compatible with CUDA $11.3$~\cite{Nickolls2008ScalablePP}), and PyTorch Lightning $2.0.0$~\cite{Falcon2019PyTorchL}.
We provide the code and additional configuration details along with this document. This includes a requirement file with all versions of the packages used and a Docker~\cite{Merkel2014DockerLL} image specification file.

\section{Detailed Architectures}
\Cref{tab:archi} depicts the detailed architecture and configuration of the networks used as perturbator and classifier in \ours{}.
Total numbers of learnable parameters are also reported; note that only the classifier is used for inference.
For the perturbator, layers 3.1 and 3.2 are both using the output of layer 2; their respective outputs are $\mu$ and $\sigma$, used to draw a new vector from a Gaussian distribution (then given to the 4\textsuperscript{th} layer), as in a classic VAE.
Regarding the classifier, embeddings used to calculate the contrastive loss are taken after layer 6.

\begin{table}[h]
    \centering
    \resizebox{\linewidth}{!}{
\setlength{\tabcolsep}{10pt} 
\renewcommand{\arraystretch}{.7}%
\begin{tabular}{|c|cc|c|c|c|c|c|}
    \hline\rule{0pt}{.9em}
    & No. & Type  & Input Size & Output Size & Affine & Bias & \# Param.\\
    \hline\hline\rule{0pt}{.9em}
    \parbox[t]{1em}{\multirow{7}{*}{\rotatebox[origin=c]{90}{Perturbator}}} & 1 & Linear & 3072 & 3072 & - & Yes & 9.4M\\
    & 2 & LeakyReLU & 3072 & 3072 & - & - & -\\
    & 3.1 & Linear & 3072 & 3072 & - & Yes & 9.4M\\
    & 3.2 & Linear & 3072 & 3072 & - & Yes & 9.4M\\
    & 4 & Linear & 3072 & 3072 & - & Yes & 9.4M\\
    & 5 & LeakyReLU & 3072 & 3072 & - & - & -\\
    & 6 & Linear & 3072 & 6144 & - & Yes & 18.8M\\
    \cline{1-8}
    \multicolumn{6}{|l}{}&\multicolumn{1}{|c}{\raisebox{-.3em}{Total}}&\raisebox{-.3em}{56.6M}\\[.2em]
    \hline\hline\rule{0pt}{.9em}
    \parbox[t]{1em}{\multirow{7}{*}{\rotatebox[origin=c]{90}{Classifier}}} & 1 & Linear & 3072 & 1024 & - & No & 3.1M\\
    & 2 & BatchNorm & 1024 & 1024 & No & - & 0\\
    & 3 & LeakyReLU & 1024 & 1024 & - & - & -\\
    & 4 & Linear & 1024 & 512 & - & No & 524K\\
    & 5 & BatchNorm & 512 & 512 & No & - & 0\\
    & 6 & LeakyReLU & 512 & 512 & - & - & -\\
    & 7 & Linear & 512 & 1 & - & No & 512\\
    \cline{1-8}
    \multicolumn{6}{|l}{}&\multicolumn{1}{|c}{\raisebox{-.3em}{Total}}&\raisebox{-.3em}{3.7M}\\[.2em]
    \hline
\end{tabular}

}
    \caption{Detailed architecture of the perturbator and classifier used in \ours{}. The layers are numbered for each network following their order of appearance in the layer sequence. When a configuration parameter (Affine, Bias...) does not apply to a layer type, a dash is used.}
    \label{tab:archi}
\end{table}

\section{Additional Results}

To study the stability of \ours{} with respect to a varying hyperparameter $\lambda$, we report per class results on CIFAR-10 with values $5$, $10$ and $20$ (\Cref{tab:lambdas-impact}).
Compared to PLAD~\cite{PLAD2022}, \ours{} shows SotA performance regardless of the value of $\lambda$ and greater stability.
In the end, we prove that \ours{} does not require any fine-tuning of the optimiser, learning rate, or other hyperparameter to achieve SotA performance.
\Cref{tab:lambdas} highlights the maximum AUC values reached for each class and $\lambda$.

\Cref{tab:cifar100-detail} depicts detailed results of \ours{} and PLAD which we also ran on CIFAR-100 (results were not provided in the original paper).
\ours{} outperforms PLAD on each meta-class and averages at 80.3\% AUC, which is higher than PLAD by 16.5pp, without any increase of learnable parameter, change in the architecture or training procedure.

\Cref{tab:fig4} shows a detailed report of the results of all recent methods on CIFAR-10, including the ones based on rotations.

\begin{table}[h]
    \centering
    \resizebox{\linewidth}{!}{

\begin{tabular}{c|c|cccccccccc|c}
  \hline
        & $\lambda$ & Plane & Car & Bird & Cat & Deer & Dog & Frog & Horse & Ship & Truck & Mean \\
  \hline\hline
\multirow{3}{*}{\rotatebox[origin=c]{90}{Max.}} & 5 &   88.0 & 87.8 & 77.1 & 79.4 & 89.7 & 79.6 & \textbf{88.4} & 86.1 & \textbf{92.6} & 92.9 & 86.16 \\ 
  & 10 &  \textbf{91.0} & \textbf{90.1 }& \textbf{80.2} & \textbf{80.8} & 89.6 & \textbf{83.7} & 84.9 & 85.8 & 89.4 & 93.8 & 86.92 \\ 
  & 20 &  89.4 & 86.5 & 75.6 & 79.0 & \textbf{90.6} & 79.8 & \textbf{88.4} & \textbf{89.5} & 88.3 & \textbf{94.0} & 86.13 \\
  \hline
  & Bests & 91.0 & 90.1 & 80.2 & 80.8 & 90.6 & 83.7 & 88.4 & 89.5 & 92.6 & 94.0 & 88.09\\
  \hline
\end{tabular}
}
    \caption{Detailed results of the maximum AUC reached for each CIFAR-10 class, over three different $\lambda$ values (5, 10, 20). The mean of those maximums are reported in the last column, and the best AUC for each class in the last row.}
    \label{tab:lambdas}
\end{table}

\begin{table}[h]
  \resizebox{\linewidth}{!}{
    
\begin{tabular}{l|c|cccccccccc|c}
    \hline &&&&&&&&&&&& \\[-1em]
        Method & $\lambda$ & Plane & Car & Bird & Cat & Deer & Dog & Frog & Horse & Ship & Truck & Mean \\
    \hline\hline &&&&&&&&&&&& \\[-1em]

    PLAD~\cite{PLAD2022} & $\sim$ & 82.5\stdv{0.4} & 80.8\stdv{0.9} & 68.8\stdv{1.2} & 65.2\stdv{1.2} & 71.6\stdv{1.1} & 71.2\stdv{1.6} & 76.4\stdv{1.9} & 73.5\stdv{1.0} & 80.6\stdv{1.8} & 80.5\stdv{0.3} & 75.1 \\
    
    PLAD\textsuperscript{$\dagger$} & 5 & 79.9\stdv{1.1} & 78.8\stdv{1.4} & 64.4\stdv{2.9} & 62.2\stdv{2.6} & 59.2\stdv{3.8} & 67.1\stdv{4.3} & 61.4\stdv{5.7} & 71.6\stdv{2.9} & 78.3\stdv{2.4} & 78.7\stdv{1.8} & 70.2\\
        \hline\hline &&&&&&&&&&&& \\[-1em]
\ours{} & 5 &  86.8\stdv{1.3} & \textbf{85.7}\stdv{1.7} & \textbf{74.5}\stdv{2.3} & 75.0\stdv{6.8} & 87.4\stdv{1.7} & 77.0\stdv{2.7} & 86.1\stdv{1.7} & 84.0\stdv{2.6} & \textbf{87.8}\stdv{3.1} & 89.6\stdv{3.2} & 83.4 \\ 

 &   10 &   \textbf{89.5}\stdv{1.0} & 85.6\stdv{2.7} & 71.7\stdv{6.0} & \textbf{78.3}\stdv{3.6} & \textbf{87.7}\stdv{1.3} & \textbf{79.5}\stdv{3.7} & 82.9\stdv{3.4} & 84.3\stdv{1.5} & 87.1\stdv{1.4} & 89.9\stdv{2.7} & 83.7 \\ 
 
 &   20 &   82.5\stdv{7.6} & 81.7\stdv{3.6} & 71.1\stdv{2.9} & 73.2\stdv{3.5} & 86.5\stdv{3.0} & 78.0\stdv{1.5} & \textbf{87.5}\stdv{0.9} & \textbf{84.6}\stdv{3.7} & 85.4\stdv{3.7} & \textbf{90.0}\stdv{2.7} & 82.0 \\ 
 \cline{3-13} &&&&&&&&&&&& \\[-1em]
   & $\sim$ & \textbf{89.5}\stdv{1.0} & \textbf{85.7}\stdv{1.7} & \textbf{74.5}\stdv{2.3} &  \textbf{78.3}\stdv{3.6} & \textbf{87.7}\stdv{1.3} &  \textbf{79.5}\stdv{3.7} & \textbf{87.5}\stdv{0.9} & \textbf{84.6}\stdv{3.7} & \textbf{87.8}\stdv{3.1} & \textbf{90.0}\stdv{2.7} & \textbf{84.5}\\

  \hline
  
    \end{tabular}

  }\\[.1cm]
  \caption{Study on the impact of $\lambda$ (CIFAR-10 dataset). Official results of the baseline PLAD~\cite{PLAD2022} are achieved with different $\lambda$ values per class ($\sim$). We reproduce the experiments (PLAD\textsuperscript{$\dagger$}) fixing $\lambda$ to the most frequent value ($\lambda=5$). \ours{} performance is also reported with $\lambda$ values fixed over classes (5,10,20), and considering the best value per class ($\sim$).}
    \label{tab:lambdas-impact}
\end{table}

\begin{table}[h]
    \centering
    \resizebox{\linewidth}{!}{
\begin{tabular}{c|c|c}
\hline
\multirow{2}{*}{Meta-Class} & \multicolumn{2}{c}{AUC}\\
\cline{2-3}
 & \textbf{\ours{}} & PLAD\textsuperscript{$\dagger$}\\
\hline\hline
0 & \textbf{75.8}\stdv{2.7} & 70.2\stdv{2.5}\\
\hline
1 & \textbf{74.9}\stdv{1.8} & 65.9\stdv{1.5}\\
\hline
2 & \textbf{88.8}\stdv{1.2} & 70.3\stdv{5.4}\\
\hline
3 & \textbf{83.2}\stdv{1.6} & 62.2\stdv{1.8}\\
\hline
4 & \textbf{84.3}\stdv{1.4} & 72.2\stdv{3.1}\\
\hline
5 & \textbf{83.5}\stdv{2.7} & 61.2\stdv{2.4}\\
\hline
6 & \textbf{88.2}\stdv{3.8} & 69.7\stdv{1.9}\\
\hline
7 & \textbf{69.8}\stdv{2.7} &  57.7\stdv{2}\\
\hline
8 & \textbf{80.1}\stdv{1.8} &  60.3\stdv{2.6}\\
\hline
9 & \textbf{78.2}\stdv{0.9} &68.8\stdv{3.8}\\
\hline
10 & \textbf{94.7}\stdv{0.4} & 70.6\stdv{4}\\
\hline
11 & \textbf{77}\stdv{1.7} &  64.5\stdv{2.6}\\
\hline
12 & \textbf{74.1}\stdv{2.4} &57.8\stdv{3.6}\\
\hline
13 & \textbf{73.6}\stdv{2.4} & 51.1\stdv{1.7}\\
\hline
14 & \textbf{86.5}\stdv{2.2} &  70.3\stdv{3.4}\\
\hline
15 & \textbf{73.1}\stdv{2.5} & 51.2\stdv{1.7}\\
\hline
16 & \textbf{65.1}\stdv{2.2} & 64.7\stdv{3.1}\\
\hline
17 & \textbf{90.1}\stdv{1.2} & 66\stdv{2.6}\\
\hline
18 & \textbf{86}\stdv{2.5} & 64.2\stdv{2}\\
\hline
19 & \textbf{78.2}\stdv{2.8} & 57.4\stdv{1.5}\\
\hline
Average & \textbf{80.3} & 63.8\\
\hline
\end{tabular}
}
    \caption{Detailed results of average AUC achieved by \ours{} and PLAD\textsuperscript{$\dagger$} (reproduced results), over 5 runs, for each meta-class of the \textbf{CIFAR-100} dataset.}
    \label{tab:cifar100-detail}
\end{table}

\begin{table*}[h]
    \centering
    \resizebox{\linewidth}{!}{
\begin{tabular}{l|l|cccccccccc|c}
        \hline
        \multicolumn{1}{c}{}&&&&&&&& \\[-1em]
        \multicolumn{2}{c}{Method} & Plane & Car & Bird & Cat & Deer & Dog & Frog & Horse & Ship & Truck & Mean \\
        \hline\hline
        &&&&&&&&&& \\[-1em]
        \multirow{10}{*}{\rotatebox{90}{Rotation Based}} & Geom \cite{Geom2018} & 74.7 & 95.7 & 78.1 & 72.4 & 87.8 & 87.8 & 83.4 & 95.5 & 93.3 & 91.3 & 86.0 \\
        & Rot$^*$ \cite{RotTrans2019} & 
        78.3\stdv{0.2} & 94.3\stdv{0.3} & 86.2\stdv{0.4} & 80.8\stdv{0.6} & 89.4\stdv{0.5} & 89.0\stdv{0.4} & 88.9\stdv{0.4} & 95.1\stdv{0.2} & 92.3\stdv{0.3} & 89.7\stdv{0.3} & 88.4 \\
        & Rot+Trans$^*$ \cite{RotTrans2019} & 
        80.4\stdv{0.3} & 96.4\stdv{0.2} & 85.9\stdv{0.3} & 81.1\stdv{0.5} & 91.3\stdv{0.3} & 89.6\stdv{0.3} & 89.9\stdv{0.3} & 95.9\stdv{0.1} & 95.0\stdv{0.1} & 92.6\stdv{0.2} & 89.8 \\
        & GOAD$^*$\cite{GOAD2020} & 75.5\stdv{0.3} & 94.1\stdv{0.3} & 81.8\stdv{0.5} & 72.0\stdv{0.3} & 83.7\stdv{0.9} & 84.4\stdv{0.3} & 82.9\stdv{0.8} & 93.9\stdv{0.3} & 92.9\stdv{0.3} & 89.5\stdv{0.2} & 85.1 \\
        & DROC~\cite{DROC2021} &90.9\stdv{0.5}&98.9\stdv{0.1}&88.1\stdv{0.1}&83.1\stdv{0.8}&89.9\stdv{1.3}&90.3\stdv{1.0}&93.5\stdv{0.6}&98.2\stdv{0.1}&96.5\stdv{0.3}&95.2\stdv{1.3}& 92.5\\
        & CSI \cite{CSI2020}  & 89.9\stdv{0.1} & 99.1\stdv{0.0} & 93.1\stdv{0.2} & 86.4\stdv{0.2} & 93.9\stdv{0.1} & 93.2\stdv{0.2} & 95.1\stdv{0.1} & 98.7\stdv{0.0} & 97.9\stdv{0.0} & 95.5\stdv{0.1} & 94.3 \\
        & iDECODe \cite{iDECODe2022} & 86.5\stdv{0.0} & 98.1\stdv{0.0} & 86.0\stdv{0.5} & 82.6\stdv{0.1} & 90.9\stdv{0.1} &89.2\stdv{0.1} & 88.2\stdv{0.4} & 97.8\stdv{0.1} &97.2\stdv{0.0}&95.5\stdv{0.1} & 91.2 \\
        & SSD \cite{SSD2021}  & 82.7 & 98.5 & 84.2 & 84.5 & 84.8 & 90.9 & 91.7 & 95.2 & 92.9 & 94.4 & 90.0 \\
        & NDA~\cite{NDA2021} & 98.5 & 76.5 & 79.6 & 79.1 & 92.4 & 71.7 & 97.5 & 69.1 & 98.5 & 75.2 & 84.3 \\            
        & UniCon-HA~\cite{UniConHA2023} & 91.7\stdv{0.1} & 99.2\stdv{0} & 93.9\stdv{0.1} & 89.5\stdv{0.2} & 95.1\stdv{0.1} & 94.1\stdv{0.2} & 96.6\stdv{0.1} & 98.9\stdv{0.0} & 98.1\stdv{0.0} & 96.6\stdv{0.1} & 95.4 \\
        & UNODE~\cite{Mirzaei_2024_CVPR} & 97.0 & 98.8 & 96.0 & 92.4 & 96.5 & 94.7 & 98.5 & 98.6 & 98.6 & 97.8 & 96.9\\
    \hline
 
        \multirow{16}{*}{\rotatebox{90}{Geometrically Unbiased}} & AnoGAN \cite{AnoGAN2017} &   67.1 & 54.7 & 52.9 & 54.5 & 65.1 & 60.3 & 58.5 & 62.5 & 75.8 & 66.5 & 61.8 \\
        & OCSVM~\cite{OCSVM2001}  & 61.6 & 63.8 & 50.0 & 55.9 & 66.0 & 62.4 & 74.7 & 62.6 & 74.9 & 75.9 & 64.7\\
        & IF~\cite{IF2008}         &   66.1 & 43.7 & 64.3 & 50.5 & 74.3 & 52.3 & 70.7 & 53.0 & 69.1 & 53.2 & 59.7 \\
        & DAE\cite{DAE2008}    &  41.1 & 47.8 & 61.6 & 56.2 & 72.8 & 51.3 & 68.8 & 49.7 & 48.7 & 37.8 & 53.6\\
        & DAGMM~\cite{DAGMM2018}      &   41.4 & 57.1 & 53.8 & 51.2 & 52.2 & 49.3 & 64.9 & 55.3 & 51.9 & 54.2 & 53.1 \\
        & ADGAN~\cite{ADGAN2019}    &  63.2 & 52.9 & 58.0 & 60.6 & 60.7 & 65.9 & 61.1 & 63.0 & 74.4 & 64.2 & 62.4\\
        & DSVDD~\cite{Ruff2018DeepOC} & 61.7 & 65.9 & 50.8 & 59.1 & 60.9 & 65.7 & 67.7 & 67.3 & 75.9 & 73.1 & 64.8\\
        & OCGAN~\cite{OCGAN2019}   &   75.7 & 53.1 & 64.0 & 62.0 & 72.3 & 62.0 & 72.3 & 57.5 & 82.0 & 55.4 & 65.6 \\
        & TQM~\cite{TQM2019} &  40.7 & 53.1 & 41.7 & 58.2 & 39.2 & 62.6 & 55.1 & 63.1 & 48.6 & 58.7 & 52.1\\
        & DROCC~\cite{DROCC2020} &  79.2 & 74.9 & 68.3 & 62.3 & 70.3 & 66.1 & 68.1 & 71.3 & 62.3 & 76.6 & 69.9\\
        & HRN-L2~\cite{HRN2020}    &  80.6 & 48.2 & 64.9 & 57.4 & 73.3 & 61.0 & 74.1 & 55.5 & 79.9 & 71.6 & 66.7\\
        & HRN~\cite{HRN2020}      &   77.3 & 69.9 & 60.6 & 64.4 & 71.5 & 67.4 & 
        77.4 & 64.9 & 82.5 & 77.3 & 71.3\\
        & DPAD~\cite{Fu2024DensePF} & 78.0\stdv{0.3} & 75.0\stdv{0.2} & 68.1\stdv{0.5} & 66.7\stdv{0.4} & 77.9\stdv{0.8} & 68.6\stdv{0.3} & 81.2\stdv{0.4} &  74.8\stdv{0.2} & 79.1\stdv{1.0} & 76.1\stdv{0.2} & 74.6\\

        & DO2HSC~\cite{Zhang2023DeepOH} & 81.3\stdv{0.2} & 82.7\stdv{0.3} & 71.3\stdv{0.4} & 71.2\stdv{1.3} & 72.9\stdv{2.1} & 72.8\stdv{0.2} & 83.0\stdv{0.6} & 75.5\stdv{0.4} & 84.4\stdv{0.5} & 82.0\stdv{0.9} & 77.7\\
        & PLAD~\cite{PLAD2022} & 82.5\stdv{0.4} & 80.8\stdv{0.9} & 68.8\stdv{1.2} & 65.2\stdv{1.2} & 71.6\stdv{1.1} & 71.2\stdv{1.6} & 76.4\stdv{1.9} & 73.5\stdv{1.0} & 80.6\stdv{1.8} & 80.5\stdv{0.3} & 75.1 \\

  & \ours{} (ours) & 89.5\stdv{1.0} & 85.7\stdv{1.7} & 74.5\stdv{2.3} &  78.3\stdv{3.6} & 87.7\stdv{1.3} &  79.5\stdv{3.7} & 87.5\stdv{0.9} & 84.6\stdv{3.7} & 87.8\stdv{3.1} & 90.0\stdv{2.7} & 84.5\\
    \hline
    \end{tabular}
}
    \caption{Detailed results (AUC) on CIFAR-10 of all methods depicted in Figure~4 of the main article. * denotes values taken from CSI~\cite{CSI2020}.}
    \label{tab:fig4}
\end{table*}

\section{Vector Space Analysis}
This section details the 3D plots in \Cref{fig:csplad-data,fig:rcsplad-data,fig:plad-feat,fig:rcsplad-feat}.
t-SNE \cite{Maaten2008VisualizingDU} was used to reduce the different vectors to 3D vectors (with a perplexity of 30), and each plot shows multiple view angles.
Models from the ablation study and features extracted from CIFAR-10 with ResNet50 are used.

\subsection{Pseudo-Anomalies}

\begin{figure}
    \centering
    \includegraphics[width=\linewidth]{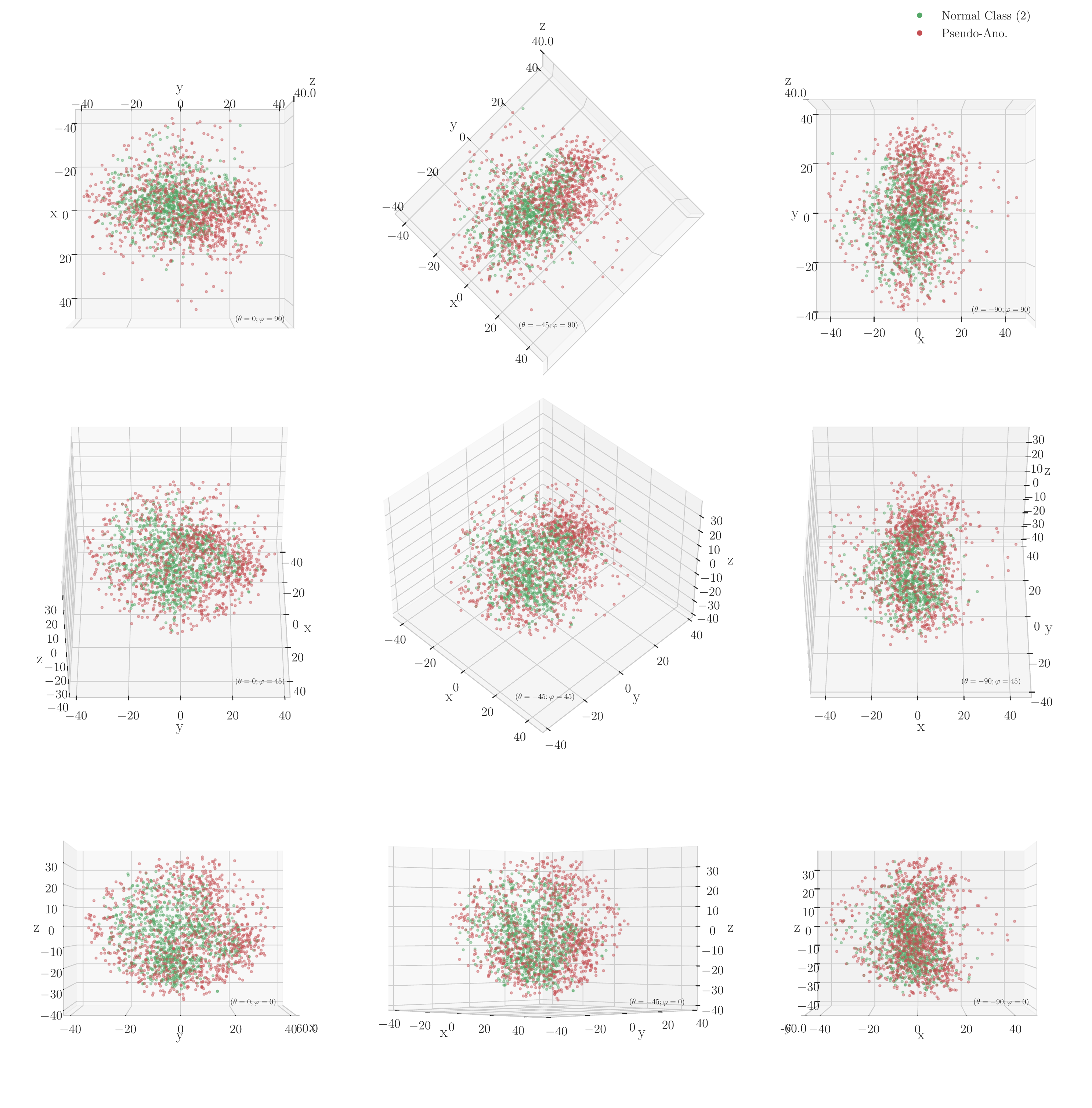}
    \caption{Study done on CIFAR-10 Bird class, with features extracted with ResNet50. Illustration of normal data (green) and generated pseudo-anomalies (red). Pseudo-anomalies were produced by the \ours{} with the AddMult perturbation method.}
    \label{fig:csplad-data}
\end{figure}

\begin{figure}
    \centering
    \includegraphics[width=\linewidth]{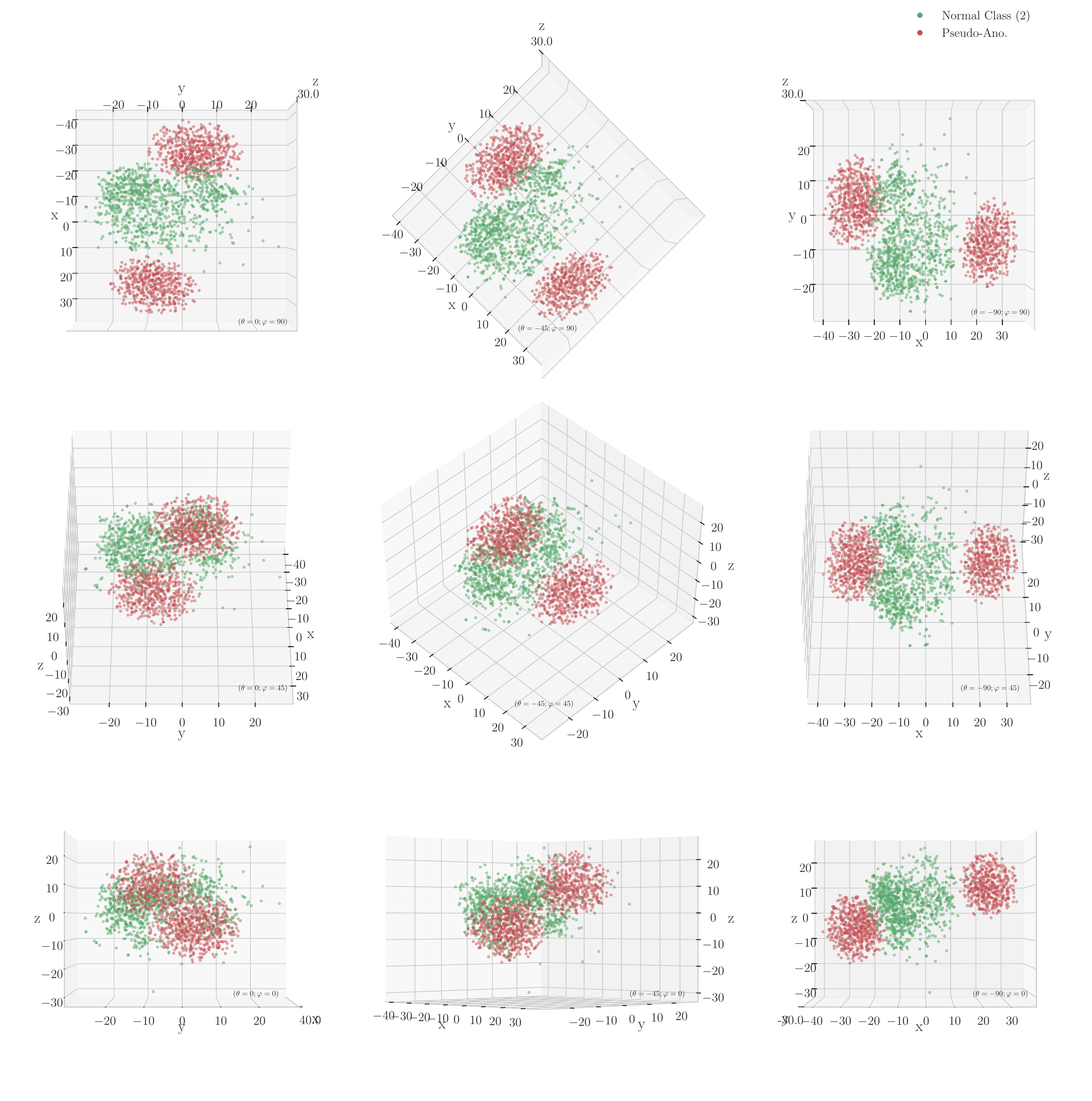}
    \caption{Study done on CIFAR-10 Bird class, with features extracted with ResNet50. Illustration of normal data (green) and generated pseudo-anomalies (red). Pseudo-anomalies were produced by a trained \ours{} configuration.}
    \label{fig:rcsplad-data}
\end{figure}

In \Cref{fig:csplad-data,fig:rcsplad-data}, normal data (green, Bird class) and the generated pseudo-anomalies (red) are displayed; features are taken before the classifier.
We compare the pseudo-anomalies generated after training with the introduced LinearMap (\ours{}, \Cref{fig:rcsplad-data}) and with the baseline mapping (\Cref{fig:csplad-data}, AddMult).
We observe that the linear mapping seems to allow for better separated features compared to the baseline noising procedure, with which normal and pseudo-anomalous features are entangled.
An interesting observation is that two ``blobs" are generated with our linear mapping (\Cref{fig:rcsplad-data}).
In fact, we found that this seems to be an amplification of the dissimilarities already present in the normal data. Indeed, by measuring the cosine similarity between the normal vectors generating one blob and the other, we observed that data of each group had greater similarities within the group than with any data from the other one.

\subsection{Embedded Vectors}

\begin{figure}
    \centering
    \includegraphics[width=\linewidth]{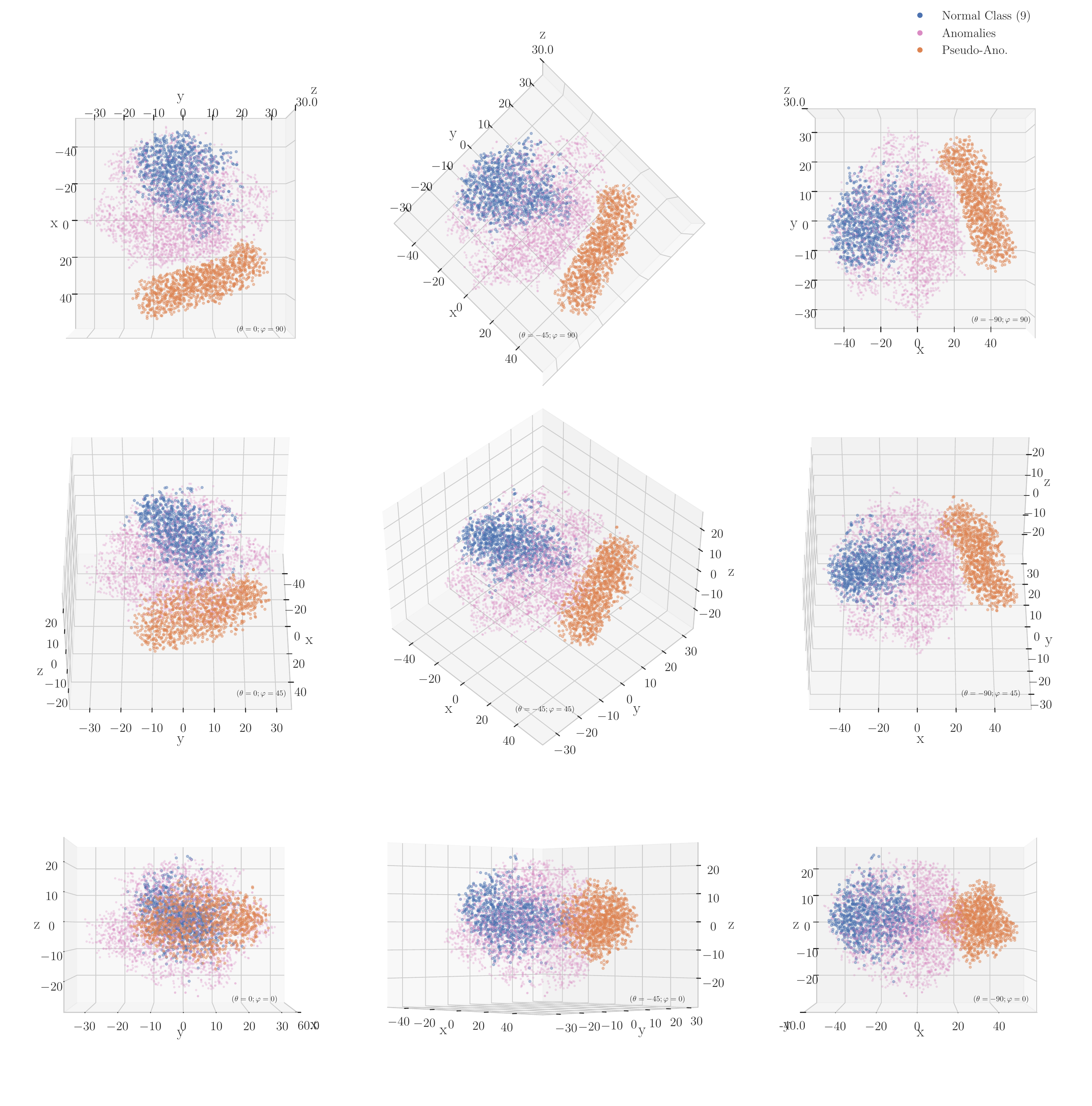}
    \caption{Study done on CIFAR-10 Truck class, with features extracted with ResNet50. Illustration of embedded vectors from normal data (blue), generated pseudo-anomalies (orange) and real anomalies (pink). Pseudo-anomalies were produced by training \ours{} with the AddMult perturbation and without contrastive loss.}
    \label{fig:plad-feat}
\end{figure}

\begin{figure}
    \centering
    \includegraphics[width=\linewidth]{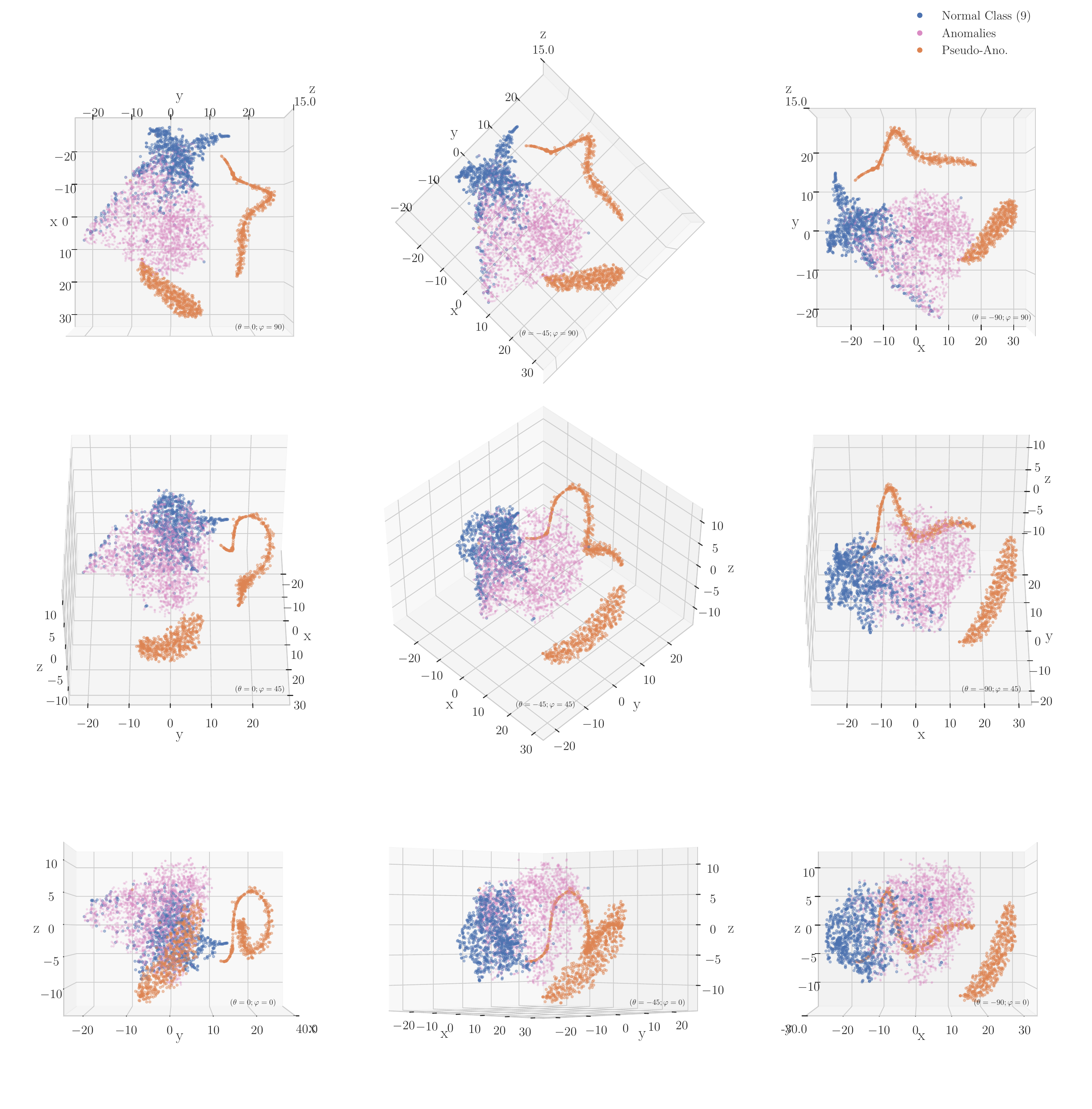}
    \caption{Study done on CIFAR-10 Truck class, with features extracted with ResNet50. Illustration of embedded vectors from normal data (blue), generated pseudo-anomalies (orange) and real anomalies (pink). Pseudo-anomalies were produced by a trained \ours{} configuration.}
    \label{fig:rcsplad-feat}
\end{figure}

In \Cref{fig:plad-feat,fig:rcsplad-feat}, embedded vectors from normal data (blue, Truck class), the generated pseudo-anomalies (orange) and real anomalies (other classes, pink) are displayed; vectors are taken after the first part of the classifier $\ffunc{f}_1$, where the contrastive loss is applied.
We compare the vectors extracted with the AddMult perturbation and without contrastive guidance (\Cref{fig:plad-feat}) and \ours{} (\Cref{fig:rcsplad-feat}).
With \ours{} we see a more important aggregation of the normal data points, and the separation with the anomalies seems clearer.

\section{Supplements on the Ablation Study}
\subsection{Mean Contrastive Loss}

In the ablation study, we report results with a simplified version of the fully contrastive loss which does not compute the similarity with respect to all vectors for each vector, but instead with respect to the mean embeddings of the normal data ($\fvm{\hat{z}} = \frac{1}{\fdim{N}}\sum\limits_{i=1}^{\fdim{N}}\fvm{z}_i$) and the pseudo-anomalies ($\fvm{\hat{\Tilde{z}}} = \frac{1}{\fdim{N}}\sum\limits_{i=1}^{\fdim{N}}\fvm{\Tilde{z}}_i$):

\begin{equation}
\ffunc{\hat{L}}_{c} = \frac{1}{\fdim{N}}\sum\limits_{i = 1}^{N}-\mathrm{log}\left[\frac{\mathrm{exp}(\ffunc{c}(\fvm{z}_i, \fvm{\hat{z}}))}{\mathrm{exp}(\ffunc{c}(\fvm{z}_i, \fvm{\hat{\Tilde{z}}})) + \mathrm{exp}(\ffunc{c}(\fvm{z}_i, \fvm{\hat{z}}))}\right]
\end{equation}

\subsection{Other Perturbations}
To supplement our ablation study we provide in \Cref{tab:ablation-supp} results with only the additive noise (Add) and only the multiplicative noise (Mult).
Both are learned with the same VAE in \ours{} with a reduced output dimension to generate only $\fvm{\alpha}_i$ or $\fvm{\beta}_i$. The results stay on par with the AddMult perturbation, though only the multiplicative part of the noise gives on average a better performance.

\begin{table*}[t]
    \centering
    \resizebox{\linewidth}{!}{
\begin{tabular}{c|c|cccccccccc|c}
  \hline
        Perturbation & Guidance & Plane & Car & Bird & Cat & Deer & Dog & Frog & Horse & Ship & Truck & Mean \\
  \hline\hline &&&&&&&&&&&&\\[-1em]
Gaussian & -  & 83.3\stdv{2.6} & 75.7\stdv{10.0} & 69.8\stdv{3.2} & 72.8\stdv{9.1} & 80.6\stdv{3.4} & 72.0\stdv{3.3} & 84.5\stdv{1.7} & 79.6\stdv{5.7} & 79.9\stdv{3.6} & 76.9\stdv{9.3} & 77.5\\

Gaussian & $\checkmark$ & 75.1\stdv{3.2} & 79.6\stdv{3.4} & 68.4\stdv{1.3} & 70.4\stdv{1.6} & 74.5\stdv{3.7} & 69.1\stdv{2.6} & 78.5\stdv{2.1} & 69.1\stdv{3.2} & 75.2\stdv{5.0} & 77.6\stdv{5.3} & 73.7 \\
\hline
AddMult & - & 66.8\stdv{3.5} & 70.5\stdv{9.5} & 67.4\stdv{1.5} & 64.9\stdv{3.6} & 67.5\stdv{5.4} & 67.0\stdv{3.5} & 70.6\stdv{7.8} & 68.1\stdv{7.7} & 72.5\stdv{4.6} & 68.7\stdv{5.2} & 68.4\\ 
AddMult & $\checkmark$ & 59.2\stdv{2.6} & 62.6\stdv{2.6} & 61.2\stdv{1.5} & 58.7\stdv{2.0} & 63.0\stdv{1.9} & 62.6\stdv{1.0} & 68.6\stdv{3.0} & 59.5\stdv{4.9} & 61.1\stdv{1.2} & 57.5\stdv{4.6} & 61.4\\ 

Add & $\checkmark$ & 59.8\stdv{2.2} & 62.7\stdv{4.4} & 59.7\stdv{2.3} & 60.1\stdv{3.3} & 59.0\stdv{2.5} & 56.6\stdv{4.3} & 61.5\stdv{7.7} & 59.7\stdv{3.2} & 59.7\stdv{4.1} & 62.0\stdv{3.5} & 60.1\\

Mult & $\checkmark$ & 62.0\stdv{3.2} & 65.0\stdv{3.7} & 62.6\stdv{3.4} & 58.5\stdv{3.0} & 61.2\stdv{2.7} & 59.4\stdv{6.1} & 66.0\stdv{5.2} & 59.2\stdv{2.9} & 66.8\stdv{3.4} & 64.5\stdv{1.0} & 62.5\\
\hline

LinearMap & - & 69.5\stdv{9.6} & 75.7\stdv{10.1} & 76.2\stdv{1.9} & 58.8\stdv{8.7} & 74.5\stdv{6.7} & 78.4\stdv{2.0} & 85.1\stdv{1.0} & 78.0\stdv{5.4} & 85.6\stdv{3.4} & 86.5\stdv{2.5} & 76.8 \\ 
LinearMap & $\checkmark$ (Mean) & 81.4\stdv{6.0} & 72.4\stdv{7.2} & 74.8\stdv{3.5} & 60.3\stdv{7.9} & 74.5\stdv{5.5} & 77.9\stdv{1.8} & 86.4\stdv{1.8} & 82.6\stdv{2.7} & 84.0\stdv{3.1} & 74.3\stdv{11.9} & 76.9\\

LinearMap & $\checkmark$ & 86.8\stdv{1.3} & 85.7\stdv{1.7} & 74.5\stdv{2.3} & 75.0\stdv{6.8} & 87.4\stdv{1.7} & 77.0\stdv{2.7} & 86.1\stdv{1.7} & 84.0\stdv{2.6} & 87.8\stdv{3.1} & 89.6\stdv{3.2} & 83.4\\ 

   \hline
\end{tabular}
}
    \caption{Ablation study of the main article supplemented by results with only additive (Add) and only multiplicative (Mult) noises.}
    \label{tab:ablation-supp}
\end{table*}

\begin{figure}
    \centering
    \resizebox{.8\linewidth}{!}{\includegraphics{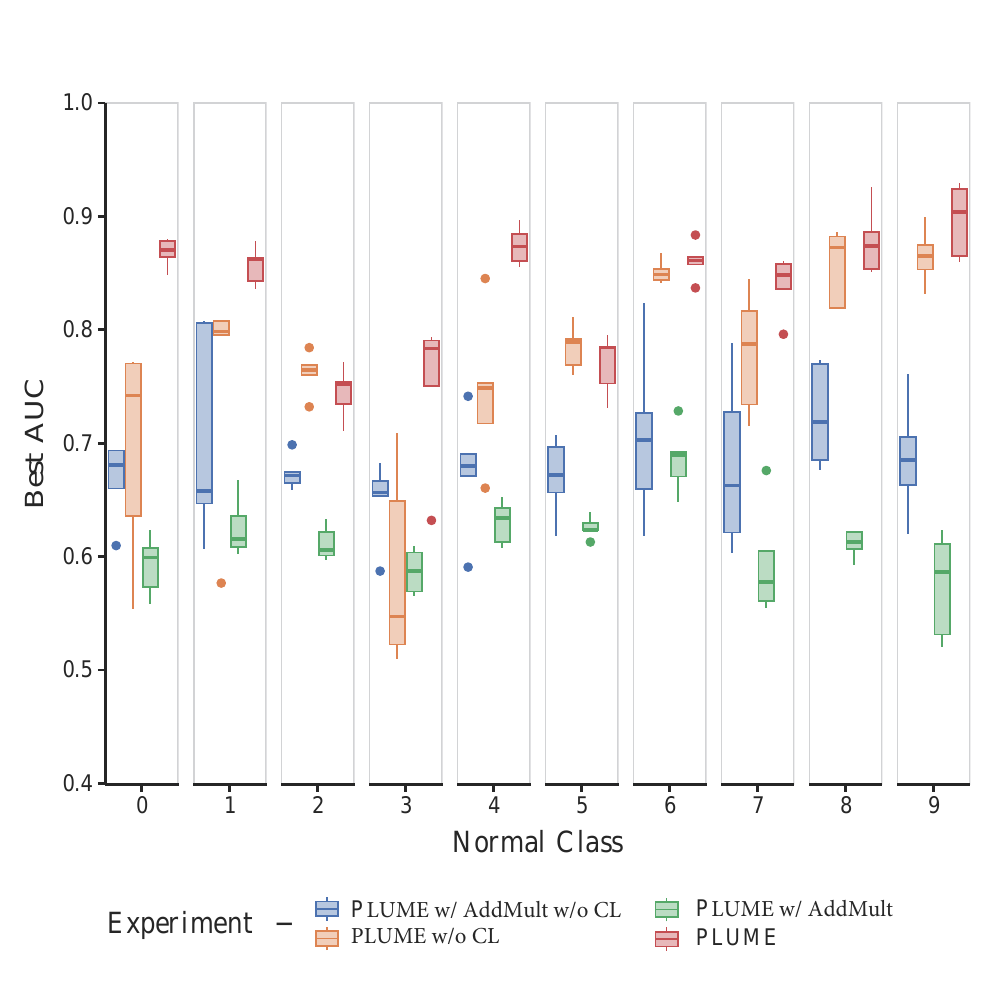}}
    \caption{Visual interpretation of the ablation study conducted in the main paper, in the form of box plots. Results are displayed by class, and coloured by experiment.}
    \label{fig:ablation}
\end{figure}

\begin{figure}
    \centering
    \includegraphics[width=2.5\linewidth,angle=90,origin=c]{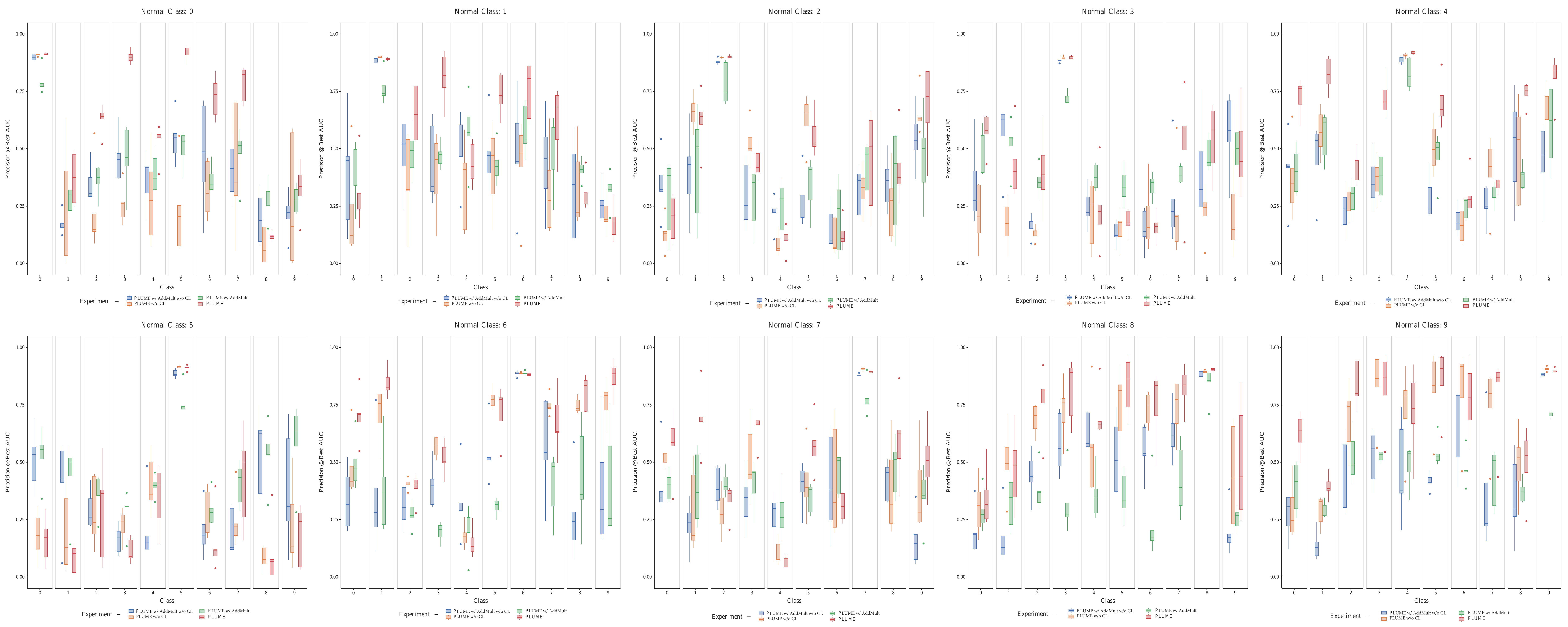}\textbf{}
    \caption{In-depth illustration of the ablation study. Each sub-figure shows the results of the trained models on one normal class. In each one, precision over the different classes are reported.}
    \label{fig:ablation-pc}
\end{figure}

\clearpage

\end{document}